\documentclass[11pt]{article}

\usepackage[final]{acl}

\usepackage{times}
\usepackage{latexsym}
\usepackage{amsmath}
\usepackage{amssymb}

\usepackage[T1]{fontenc}

\usepackage[utf8]{inputenc}

\usepackage{microtype}

\usepackage{inconsolata}

\usepackage{graphicx}
\usepackage[most]{tcolorbox}
\usepackage{listings}
\usepackage{xcolor}
\usepackage{booktabs}
\usepackage{arydshln}
\usepackage{algorithm}
\usepackage{algpseudocode}

\title{A Shared Future-Attention Direction in Key Space for KV Cache Eviction}

\author{
  Hikaru Shijo \and Yutaka Yoshihama \\
  Panasonic Automotive Systems Co., Ltd. \\
  \texttt{shijo.hikaru@jp.panasonic.com}, \texttt{yoshihama.yutaka@jp.panasonic.com}
}

\begin{document}
\maketitle
\begin{abstract}
Long-context inference is increasingly constrained by the memory and computation
of the key--value (KV) cache.
We show that cumulative future attention can be accurately approximated by a
single linear direction in key space.
After correcting for RoPE-induced rotations, this direction exhibits a
dominant component shared across inputs within each model, layer, and KV head.
Based on this finding, we propose \textsc{FAS-KV}, which learns a shared
future-attention direction offline and scores cached tokens using a single
key-space projection at runtime.
At inference time, the method requires neither future queries nor access to
attention matrices, making it lightweight and compatible with FlashAttention.
Across LongBench, RULER, and Needle-in-a-Haystack, \textsc{FAS-KV} achieves
the best or near-best performance among eviction methods,
while maintaining low scoring overhead.
\end{abstract}
\begingroup
\makeatletter
\renewcommand{\@makefntext}[1]{\noindent#1}
\makeatother
\renewcommand{\thefootnote}{}
\footnotetext{\mbox{Code is available at \href{https://github.com/shijo-hikaru/FAS-KV}{github.com/shijo-hikaru/FAS-KV}.}}
\endgroup

\section{Introduction}
\label{sec:introduction}

Large language models (LLMs) increasingly rely on long contexts, but their
key--value (KV) caches grow linearly with context length and batch size,
creating a major memory and bandwidth bottleneck. KV-cache eviction reduces
this cost by retaining only tokens expected to remain useful.

\begin{figure}[t]
  \centering
  \includegraphics[width=0.9\linewidth]{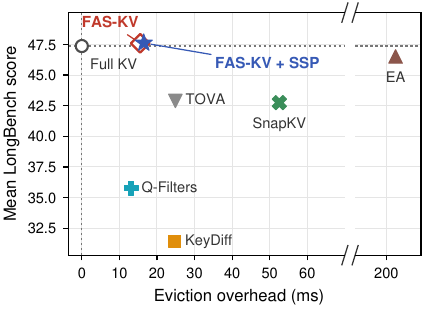}
  \caption{Accuracy--efficiency trade-off for Qwen3-8B after
evicting 70\% of the context tokens from the KV cache.
The vertical axis reports mean LongBench score; the
horizontal axis reports prefill overhead over 50 repeated runs of one natural
24K-token context on an NVIDIA H200 GPU (batch size 1). FAS-KV maintains near-full-cache
performance with low scoring overhead.
  }

  \label{fig:intro_motivation}
\end{figure}

Existing methods commonly estimate token importance from observed attention
scores, as in H$_2$O, TOVA, and SnapKV
\citep{zhang2023h2o,oren2024tova,li2024snapkv}.
However, fused kernels such as FlashAttention do not materialize the full
attention matrix, so accessing token-level statistics requires custom kernel
support or a less efficient implementation
\citep{dao2022flashattention,dao2023flashattention2}.
Attention-free alternatives include Q-Filters, which scores keys using a
context-agnostic direction from query geometry, and KeyDiff, which uses
within-cache geometry \citep{godey2025qfilters,park2025keydiff}. Expected
Attention (EA) models future queries with a Gaussian distribution and computes
their expected attention \citep{devoto2025expectedattention}. EA achieves
strong accuracy but incurs substantial computational overhead, as shown in
Figure~\ref{fig:intro_motivation}. 
More generally, query-head-wise scoring is less naturally aligned with grouped-query attention (GQA), where multiple query heads share each KV head: producing one eviction score requires aggregation across the group \citep{ainslie2023gqa}.

This motivates scoring directly at the KV-head level. For eviction, we need
not reconstruct individual future queries; we only need to rank cached tokens
by the cumulative attention they are expected to receive. We therefore ask
whether this quantity can be predicted directly from cached keys, without
future queries or attention matrices. Centered log-attention from a single
query is exactly linear in key space because centering cancels the softmax
normalizer. More importantly, we find that cumulative attention from multiple
future queries is accurately approximated by a single direction. After
compensating for the boundary-dependent rotation
introduced by rotary positional embeddings (RoPE), the fitted directions
exhibit a dominant component shared across inputs within each model, layer,
and KV head.

Based on this finding, we propose \textsc{FAS-KV}, a lightweight, query-free
KV-cache eviction method. Offline, we align fitted future-attention directions
in a boundary-relative coordinate system, form their normalized direct mean,
and fit its scale. At inference time, the shared direction is transported to
the current boundary and scores each cached token with one key--direction inner
product. The method requires neither future queries nor attention matrices and
remains compatible with FlashAttention.

Because token-wise eviction can fragment lexical units split into multiple
subwords, we additionally introduce Selective Span Propagation (SSP).
SSP propagates scores within tokenizer-defined spans at a small calibrated
set of layers, improving robustness without changing the cache budget or
requiring attention matrices.

We evaluate \textsc{FAS-KV} on LongBench, RULER, and Needle-in-a-Haystack across
multiple LLM families and compression ratios. It preserves strong long-context
and retrieval performance under aggressive compression with minimal scoring
overhead. On LongBench with Qwen3-8B, it reduces the KV-cache size by 70\% while
maintaining the full-cache average.
Our contributions are summarized as follows:

\begin{itemize}
  \setlength{\topsep}{0pt}
  \setlength{\itemsep}{0cm} 
  \setlength{\parskip}{0cm} 

    \item We identify a shared canonical future-attention direction for each
    model, layer, and KV head.

    \item We use this direction to score cached tokens with one projection per key,
    without future queries or attention matrices.

    \item Experiments across models and long-context benchmarks demonstrate
effective compression with low measured overhead, while SSP further improves
retrieval robustness under aggressive compression.
\end{itemize}

\section{Preliminaries and Problem Formulation}

\paragraph{Attention and KV caching.}
Let $h_i \in \mathbb{R}^{d_{\mathrm{model}}}$ denote the hidden state at position $i$.
For a single attention head, the query, key, and value vectors are given by
\begin{equation}
\begin{aligned}
q_i &= R_i W_Q h_i, \quad
k_i = R_i W_K h_i, \quad
v_i = W_V h_i.
\end{aligned}
\end{equation}
Here, $R_i$ denotes the RoPE rotation at position $i$~\citep{su2024roformer}.
At decoding step $t$, the attention weight assigned to a cached token $i\leq t$ is
\begin{equation}
a_{t,i}
=
\frac{\exp\!\left(q_t^\top k_i/\sqrt{d}\right)}
{\sum_{j\leq t}\exp\!\left(q_t^\top k_j/\sqrt{d}\right)}.
\label{eq:attention_weight}
\end{equation}
Autoregressive inference stores all preceding keys and values to avoid repeated
projection of past hidden states. Consequently, both KV-cache memory and the
cost of attending to the cache grow linearly with the context length.

\paragraph{Direct contribution to future attention outputs.}
For one attention head before the output projection, the output at a future
position $t$ is $o_t=\sum_{j\leq t}a_{t,j}v_j$.
The direct contribution of a past token $i$ and its magnitude are
\begin{equation}
\Delta o_{t,i}=a_{t,i}v_i,
\qquad
\lVert\Delta o_{t,i}\rVert_2=a_{t,i}\lVert v_i\rVert_2.
\label{eq:direct_token_contribution}
\end{equation}

The norm equality is exact because attention weights are nonnegative.
For a future window $\mathcal{F}$ of length $H=|\mathcal{F}|$, define the
average future attention and average direct contribution by
\begin{subequations}
\label{eq:future_contribution}
\begin{align}
\alpha_i^{\mathrm{future}}
&=
\frac{1}{H}\sum_{t\in\mathcal{F}}a_{t,i},
\label{eq:future_attention_mass}
\\
D_i^{\mathrm{future}}
&=
\frac{1}{H}\sum_{t\in\mathcal{F}}
\lVert\Delta o_{t,i}\rVert_2
=
\alpha_i^{\mathrm{future}}\lVert v_i\rVert_2.
\label{eq:future_contribution_factorization}
\end{align}
\end{subequations}

Thus, average future direct contribution factorizes exactly into average future
attention and the value norm. We omit the superscript ``future'' when the
observation window is clear from context.
This oracle measures the magnitude of a token's direct contribution to the attention output, rather than the exact impact of token removal (see Appendix~\ref{app:deletion_effect}).

\paragraph{Future importance under a cache budget.}
Consider an eviction boundary $b$, at which only $K<b$ cached tokens can be
retained. For the horizon
$\mathcal{F}_b=\{b+1,\ldots,b+H\}$, we use
$D_i^{\mathrm{future}}$ as the oracle importance of token $i\leq b$.
An optimal direct-contribution policy would retain the tokens with the largest
$D_i^{\mathrm{future}}$. This oracle, however, is unavailable at the eviction
boundary because the future queries have not yet been generated.
Existing methods therefore rely on observable proxies, such as past attention
statistics, recent queries, or pairwise key relations. We instead ask whether
the cumulative-attention factor $\alpha_i^{\mathrm{future}}$ has a reusable
geometric structure that can be inferred from the cached keys themselves.
In the following section, we show that centered log attention is linear in
centered key representations and that the canonicalized predictive directions
contain a dominant component shared across inputs. Combined with the exact value-norm
factor in Eq.~\eqref{eq:future_contribution_factorization}, this observation
enables future-contribution scoring without materializing attention matrices at
inference time.


\begin{figure}[t]
    \centering
    \includegraphics[width=\linewidth]{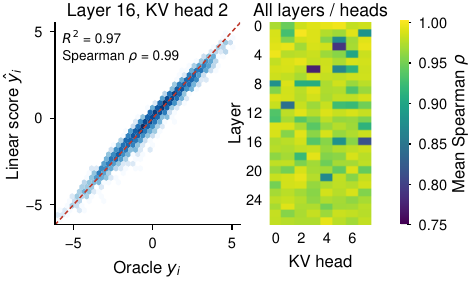}
    \caption{Within-sample prediction of cumulative
    future attention for Qwen3-1.7B on WikiText~\citep{merity2016pointer}.
    For each sample, we fit a separate direction $w^\star$ for each layer and
    KV head. \textbf{Left:} Each point represents one cached token from one
    sample at layer 16, KV head 2. The axes show the oracle centered
    log-attention $y_i$ and its fitted value $\hat{y}_i$.
    \textbf{Right:} For each layer and KV head, we compute Spearman's $\rho$
    between fitted scores and cumulative future attention across tokens within
    each sample, then average $\rho$ across samples.}
    \label{fig:linear_approx}
\end{figure}

\begin{figure*}[t]
    \centering
    \includegraphics[width=\textwidth]{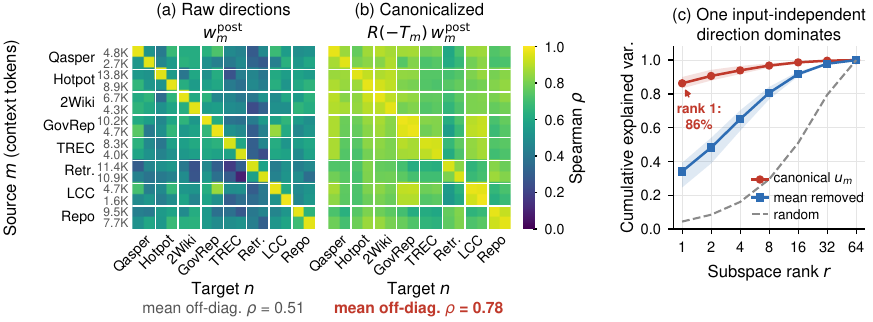}
    \caption{Boundary alignment reveals a transferable shared component in
    future-attention directions for Llama-3.2-1B-Instruct. \textbf{(a)} Cross-sample transfer before
    boundary canonicalization. \textbf{(b)} Transfer after canonicalization,
    which substantially increases rank correlation across samples and tasks.
    \textbf{(c)} Cumulative explained variance of the canonical directions and
    the residual after removing the shared mean direction.}
    \label{fig:w_transfer}
\end{figure*}

\section{Future Attention as a Shared Linear Functional}
\label{sec:future_attention_geometry}
Before introducing our eviction method, we examine the geometry of token
importance under future attention. For a single query, centered log-attention
is exactly linear in centered key space. We further show empirically that
importance aggregated over multiple future queries is well approximated by a
single direction, which becomes transferable across inputs after boundary
canonicalization.

\subsection{Exact Linearity for a Single Future Query}
\label{sec:single_query_linearity}

Consider a future query $q_t\in\mathbb{R}^{d}$ and $N$ cached keys
$\{k_i\}_{i=1}^{N}$ for a single attention head. Let $a_{t,i}$ denote the
corresponding attention weight defined in Eq.~\eqref{eq:attention_weight}, and
let $\bar{k}=N^{-1}\sum_{j=1}^{N}k_j$ be the mean cached key. Define the
centered log-attention as
\begin{subequations}
\begin{align}
\widetilde{a}_{t,i}
&=
\log a_{t,i}
-
\frac{1}{N}
\sum_{j=1}^{N}
\log a_{t,j},
\label{eq:centered_log_attention}
\\
&=
(k_i-\bar{k})^\top\frac{q_t}{\sqrt{d}}.
\label{eq:exact_log_attention}
\end{align}
\end{subequations}

Centering removes the softmax normalizer while preserving the attention
ranking. Therefore, Eq.~\eqref{eq:exact_log_attention} shows that a fixed query
ranks cached tokens exactly by projecting the centered keys onto
$q_t/\sqrt{d}$. This identity serves as
theoretical motivation; our central claim concerns cumulative attention from
multiple future queries. Appendix~\ref{app:exact_linearity} provides the full
derivation.

\subsection{Cumulative Future-Attention Direction}
\label{sec:cumulative_linearity}

For a single input with future observation window $\mathcal{F}$, we use the
average future attention $\alpha_i\equiv\alpha_i^{\mathrm{future}}$
(Eq.~\eqref{eq:future_attention_mass}) as the importance of cached token $i$.
This average normalizes for the window length while preserving the token
ranking induced by cumulative attention for a fixed $\mathcal{F}$. We use its
centered logarithm as the regression target:
\begin{equation}
y_i
=
\log(\alpha_i+\delta)
-
\frac{1}{N}
\sum_{j=1}^{N}
\log(\alpha_j+\delta).
\label{eq:centered_future_importance}
\end{equation}

We use $\delta=10^{-6}$ to avoid taking the logarithm of a numerically zero
value.
Unlike the single-query identity, $\alpha_i$ averages attention distributions
produced by different queries and different softmax normalizers. Therefore,
$y_i$ is not generally linear in $k_i$ and cannot be represented exactly by
the mean future query. Nevertheless, for each input, we estimate an
input-specific regularized linear direction $w^\star$ by regressing $y_i$ on
the centered keys, such that
\begin{equation}
y_i
\approx
(k_i-\bar{k})^\top w^\star,
\label{eq:cumulative_linear_model}
\end{equation}

where $w^\star$ summarizes the combined effect of the future queries for that
input. Rather than reconstructing individual future queries, it captures each
key's aggregate responsiveness to the future-query window.
The approximate linearity is explained by the weighted covariance of future-query directions, which governs the curvature of log cumulative attention. Small curvature along key displacements yields a small linearization error without requiring collinear queries (Appendix~\ref{app:local_linearization}).
The left panel of Figure~\ref{fig:linear_approx} shows that one direction captures most of
the variation in the centered log cumulative attention.
The right panel of Figure~\ref{fig:linear_approx} shows high Spearman correlations
between sample-specific linear scores and oracle cumulative future attention
across most layers and KV heads, indicating that a single fitted direction can
effectively rank cached tokens within each input.

\subsection{Input-Independent Direction after Boundary Canonicalization}
\label{sec:shared_canonical_direction}

The direction in Eq.~\eqref{eq:cumulative_linear_model} is fitted to post-RoPE
keys and is therefore expressed in the absolute positional frame of its input.
For calibration sample $m$, let $T_m$ denote the scalar position of its context--future boundary, and let $R(t)$ denote the RoPE rotation matrix at position $t$.
Let $\widehat{w}_m^{\mathrm{post}}$ be the normalized direction fitted for
input $m$, whose context--future boundary is $T_m$. Directions fitted at
different boundaries contain different RoPE rotations even when their
underlying predictive geometry is similar. Consequently, directly applying a
source direction to another input gives weak and variable off-diagonal
correlations, as shown in Figure~\ref{fig:w_transfer}(a).

We remove this boundary-dependent rotation by mapping each direction to the
canonical frame,
$u_m=R(-T_m)\widehat{w}_m^{\mathrm{post}}$. To evaluate transfer to a target
input $n$, we rotate it into the target frame as $R(T_n)u_m$ and use it to
score the target keys. Figure~\ref{fig:w_transfer}(b) shows that this alignment
substantially improves cross-input rank correlation across tasks and context
lengths. The correction aligns the fitted directions at their eviction
boundaries; it does not individually undo the positional rotation of every
cached key.

Figure~\ref{fig:w_transfer}(c) further shows that rank one explains $86\%$ of
the variance among the aligned directions. Thus, after boundary
canonicalization, a single input-independent vector for each layer and KV head
captures how strongly cached keys respond to future queries and thereby
predicts their future importance.

\section{Method}
\label{sec:method}

Motivated by the input-independent component identified in
Section~\ref{sec:shared_canonical_direction}, FAS-KV replaces the unavailable
input-specific future-attention direction with one offline-calibrated
canonical direction for each layer and KV head.

\subsection{Offline Calibration}
\label{sec:offline_calibration}
\label{sec:oracle_direction}

For each calibration sequence $m$, we reserve up to 128 continuation tokens as
an oracle future window. We construct the centered log-attention target from
Eq.~\eqref{eq:centered_future_importance} and fit an input-specific,
regularized linear direction $w_{m,\ell,h}^{\star,\mathrm{post}}$ from the
centered post-RoPE keys. For GQA models, attention is first averaged over all
query heads sharing a KV head, yielding one regression target per KV head.
Permanently retained sink tokens are excluded from the fit.
We calibrate once per model checkpoint using 192 prose documents, without downstream evaluation examples, and reuse the resulting directions across tasks. Calibration data and robustness analyses are detailed in Appendix~\ref{app:calibration_data}.

Since $w_{m,\ell,h}^{\star,\mathrm{post}}$ is expressed in the post-RoPE frame at boundary $T_m$, we normalize and map it to the canonical frame:
\begin{equation}
u_{m,\ell,h}
=
R(-T_m)
\frac{w_{m,\ell,h}^{\star,\mathrm{post}}}
{\lVert w_{m,\ell,h}^{\star,\mathrm{post}}\rVert_2}.
\label{eq:boundary_alignment}
\end{equation}

This common rotation preserves the fitted scores while aligning samples by
relative position. We estimate the shared orientation using the normalized
direct mean:
\begin{equation}
\mu_{\ell,h}
=
\operatorname{normalize}
\left(
\frac{1}{M}\sum_{m=1}^{M}u_{m,\ell,h}
\right).
\label{eq:canonical_directional_mean}
\end{equation}

Direction normalization removes the magnitude of each oracle fit. To restore
this scale, we fit a one-dimensional ridge coefficient for each calibration
sample and average it across samples:
\begin{equation}
\beta_{\ell,h}^{\mathrm{fit}}
=
\frac{1}{M}
\sum_{m=1}^{M}
\frac{z_{m,\ell,h}^{\top}y_{m,\ell,h}}
{z_{m,\ell,h}^{\top}z_{m,\ell,h}+\lambda_\beta}.
\label{eq:fitted_scale}
\end{equation}

where $z_{m,\ell,h}$ collects the scores obtained by transporting
$\mu_{\ell,h}$ to boundary $T_m$ and projecting the centered calibration keys;
we use $\lambda_\beta=10^{-2}$. We then store
\begin{equation}
w_{\ell,h}^{\mathrm{FAS,can}}
=
\beta_{\ell,h}^{\mathrm{fit}}\mu_{\ell,h}.
\label{eq:fitted_canonical_direction}
\end{equation}

For inference, we retain only this scaled canonical
vector per layer and KV head. Implementation
details, including GQA aggregation, calibration-corpus construction, ridge
regularization, boundary canonicalization, and the scalar fit, are provided in
Appendices~\ref{app:gqa}--\ref{app:beta_fitting}.

\subsection{Runtime Scoring and Eviction}
\label{sec:runtime_scoring}

At an eviction boundary $T$, the shared canonical direction is transported to
the current post-RoPE frame once:
\begin{equation}
w_{\ell,h}^{\mathrm{FAS}}(T)
=
R(T)w_{\ell,h}^{\mathrm{FAS,can}}.
\label{eq:runtime_fas_direction}
\end{equation}

Each cached token is assigned the logit
\begin{equation}
\widehat{a}_{\ell,h,i}
=
\left(k_{\ell,h,i}^{\mathrm{post}}\right)^\top
w_{\ell,h}^{\mathrm{FAS}}(T).
\label{eq:key_logit}
\end{equation}

The centered-key form in Section~\ref{sec:cumulative_linearity} is equivalent:
the subtracted mean contributes the same offset to every token and therefore
cancels in the Top-$K$ ranking. The exact factorization in
Eq.~\eqref{eq:future_contribution_factorization} applies to oracle future
attention, whereas $\widehat{a}_{\ell,h,i}$ only predicts its centered
logarithm. Motivated by this factorization, we define the runtime score as
follows, omitting layer and KV-head indices.
\begin{equation}
\boxed{
s_i
=
\gamma\widehat{a}_i+\log(\lVert v_i\rVert_2+\epsilon)
}
\label{eq:value_aware_score}
\end{equation}

Here, $\gamma$ controls the weight of the approximate log-attention
score relative to the log value-norm term. Unless otherwise specified, we use $\gamma=0.1$.
We set $\epsilon=10^{-6}$ for numerical stability. This log-space
formulation avoids explicitly computing a softmax.

We permanently retain the first $K_{\mathrm{sink}}=4$ sink tokens
and select the remaining $K-K_{\mathrm{sink}}$ tokens by Top-K score,
where $K$ is the total retained-cache budget.
The complete selection procedure is given in
Appendix~\ref{app:eviction}.

\subsection{Lightweight Scoring}
\label{sec:lightweight_scoring}

FAS-KV stores only one $d$-dimensional canonical vector per layer and KV head.
At each eviction boundary, rotating this vector costs
$\mathcal{O}(d)$. Scoring $T$ cached tokens takes $\mathcal{O}(Td)$ time per
layer and KV head, including the key projections and the log value-norm term.
The additional storage is $\mathcal{O}(d)$ per layer
and KV head.

Unlike attention-based or future-query estimation methods, FAS-KV requires no
attention matrices, query sampling, pairwise key comparisons, or repeated
projections onto multiple directions. Its scorer can therefore be applied
outside the attention kernel while retaining FlashAttention compatibility. We
empirically evaluate runtime overhead in Section~\ref{sec:efficiency}.

\begin{figure*}[t!]
  \centering
  \includegraphics[width=1.0\textwidth]{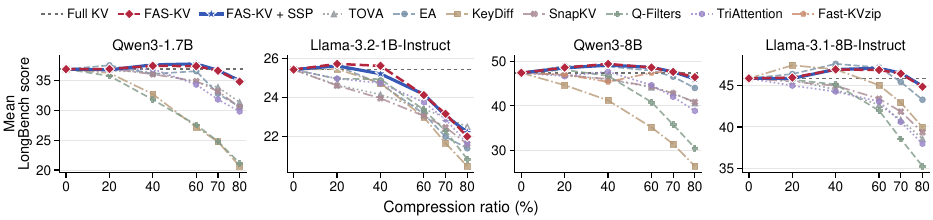}
  \caption{Mean LongBench score under increasing KV-cache compression for
  Qwen3-1.7B, Llama-3.2-1B-Instruct, Qwen3-8B, and Llama-3.1-8B-Instruct. The
  horizontal dotted line marks full-cache performance. FAS-KV and FAS-KV +
  SSP remain close to this reference across models, including under aggressive
  compression.
  }
  \label{fig:longbench_results}
\end{figure*}

\begin{figure*}[t]
  \centering
  \includegraphics[width=\textwidth]{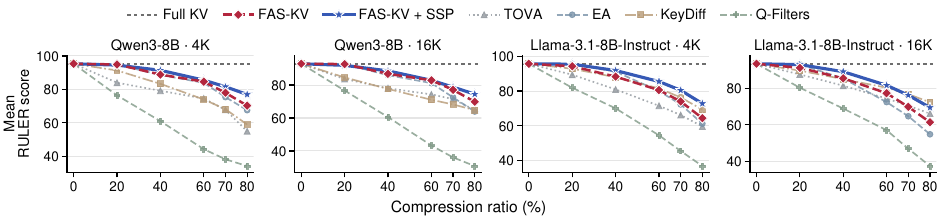}
  \caption{Mean RULER score under increasing KV-cache compression for Qwen3-8B
  and Llama-3.1-8B-Instruct at 4K and 16K context lengths. The horizontal
  dotted line marks full-cache performance. FAS-KV degrades gradually, while
  SSP provides additional gains at high compression.
  }
  \label{fig:ruler_results}
\end{figure*}

\subsection{Selective Span Propagation (SSP)}
\label{sec:optional_span_propagation}

Independent Top-$K$ selection can keep only some of the subword pieces that form a lexical unit, disrupting the information represented by the complete span. SSP addresses this structural failure mode as an optional post-scoring operation.
As an auxiliary safeguard against subword fragmentation, we identify a sparse
layer set $\mathcal{L}_{\mathrm{span}}$ using one short synthetic retrieval
sample. We first identify a responsive contiguous layer range and then retain
the smallest subset achieving the best retrieval score. Let $G(i)$ be the
tokenizer-defined lexical span containing token $i$. At the selected layers,
we propagate the largest score within each span:
\begin{equation}
\widetilde{s}_{\ell,h,i}
=
\begin{cases}
\displaystyle\max_{j\in G(i)}s_{\ell,h,j},
& \ell\in\mathcal{L}_{\mathrm{span}},\\
s_{\ell,h,i}, & \ell\notin\mathcal{L}_{\mathrm{span}}.
\end{cases}
\label{eq:main_span_score_propagation}
\end{equation}

This optional propagation encourages joint subword retention without changing
the cache budget or requiring additional attention computation. Full span
construction and layer-selection details are given in
Appendix~\ref{sec:layer_selective_span}.

\section{Experiments}

\subsection{Experimental Setup}

\paragraph{Benchmarks.}
We evaluate FAS-KV on three complementary long-context benchmarks:
LongBench~\citep{bai2024longbench}, RULER~\citep{hsieh2024ruler}, and Needle-in-a-Haystack (NIAH)~\citep{kamradt2023niah}.
LongBench evaluates performance across a diverse collection of long-context
language understanding tasks. We also evaluate on
LongBench-V2~\citep{bai2025longbenchv2}, a newer benchmark with challenging
multiple-choice questions about long contexts.
RULER provides a controlled evaluation of long-context reasoning and
retrieval; we report results at context lengths of 4K and 16K tokens. NIAH
further measures whether information embedded at different positions in a
long context can be retrieved after KV-cache compression.

\paragraph{Evaluation protocol.}
LongBench, LongBench-V2, RULER, and NIAH use greedy decoding with a single run
per configuration. We compress the prefilled context before appending the
question, ensuring query-agnostic eviction. Implementation, hardware, and
compression details are provided in Appendix~\ref{app:hardware_compression}.

\paragraph{Models.}
To evaluate FAS-KV across model families and scales,
we use Llama-3.2-1B-Instruct~\citep{meta2024llama32}, Qwen3-1.7B,
Llama-3.1-8B-Instruct~\citep{grattafiori2024llama3herdmodels}, and
Qwen3-8B~\citep{yang2025qwen3technicalreport} in our main experiments.

\begin{figure*}[t!]
  \centering
  \includegraphics[width=0.95\textwidth]{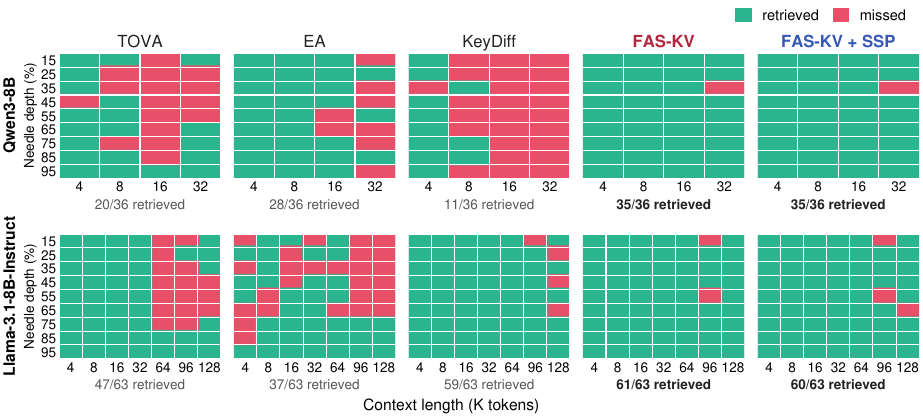}
  \caption{Needle-in-a-Haystack retrieval accuracy at 70\% KV-cache
  compression across context lengths and needle depths. The top row evaluates Qwen3-8B up
  to 32K tokens, and the bottom row evaluates Llama-3.1-8B up to 128K tokens.
  Green and red cells denote successful and unsuccessful retrieval,
  respectively.}
  \label{fig:niah_results}
\end{figure*}

\subsection{Main Results}

\paragraph{LongBench.}
Figure~\ref{fig:longbench_results} reports the mean LongBench score for four
models. FAS-KV stays near the full-cache reference through moderate compression
and degrades more gradually at the highest ratios. For both 8B models, it
matches or exceeds the full-cache score through 70\% compression, with FAS-KV
+ SSP showing similar behavior. At 80\% compression, Llama-3.2-1B-Instruct
shows a larger drop, whereas Qwen3-1.7B remains robust; both variants remain
competitive. Baseline behavior is more model
dependent: EA is often the strongest alternative, whereas KeyDiff drops
sharply on the Qwen models and Q-Filters degrades rapidly on
Llama-3.1-8B-Instruct. Overall, the core FAS-KV scorer provides a consistent
compression curve, and LongBench scores are generally similar with and without SSP.
See Figure~\ref{fig:longbench_bootstrap_diff} for bootstrap intervals on score differences.

\paragraph{RULER.}
Figure~\ref{fig:ruler_results} shows the same gradual degradation on RULER as
both context length and compression increase. FAS-KV remains competitive and adding SSP yields the strongest curve for
Qwen3-8B at both 4K and 16K and for Llama-3.1-8B-Instruct at 4K, with the
largest gains appearing at 70--80\% compression. At 16K on
Llama-3.1-8B-Instruct, FAS-KV + SSP leads through most compression ratios,
although KeyDiff is slightly higher at 80\%. EA remains competitive at
moderate ratios but drops more sharply at high compression, while Q-Filters
degrades substantially on both the Qwen and Llama models. These results separate a strong core scorer from the additional retrieval benefit provided by SSP.
See Figure~\ref{fig:ruler_bootstrap_diff} for the corresponding RULER intervals.

\paragraph{Needle-in-a-Haystack.}
Figure~\ref{fig:niah_results} evaluates retrieval across both context length
and needle depth. FAS-KV produces nearly complete success grids for Qwen3-8B
through 32K and for Llama-3.1-8B through 128K across the tested insertion depths. FAS-KV + SSP shows
similarly strong coverage.

\begin{figure*}[t]
  \centering
  \includegraphics[width=\textwidth]{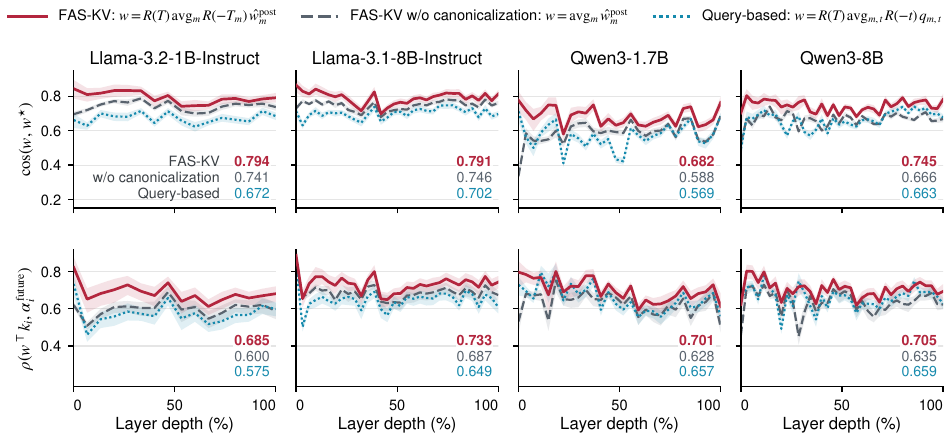}
  \caption{Held-out validation on 160 LongBench examples per model. Top:
  cosine similarity to sample-specific oracle directions; bottom: Spearman
  correlation between predicted rankings and realized cumulative attention.
  Baselines omit canonicalization or use the mean future query. Curves show
  layer means with 95\% hierarchical-bootstrap confidence intervals.}
  \label{fig:shared_direction_heldout}
\end{figure*}

\paragraph{Additional comparisons.}
On LongBench-V2 with Qwen3-8B, FAS-KV achieves 32.41\% accuracy at 70\% eviction, within one percentage point of Full KV (33.40\%). FAS-KV + SSP reaches 34.19\%, the highest accuracy in Table~\ref{tab:longbench_v2_comparison}.

The Q-Filters baseline in Figure~\ref{fig:longbench_results} is based on its official implementation. To compare the learned directions on LongBench, we use the same linear scorer for both methods, omit the value-norm term, and match the calibration data, Sink-4 protection, and per-head cache budgets. Under these conditions, the FAS-KV direction scores 4.03--10.95 points higher than the Q-Filters direction across both models and both eviction rates (Table~\ref{tab:longbench_matched_directions}).

\begin{table}[t]
\centering
\small
\setlength{\tabcolsep}{3pt}
\begin{tabular}{@{}lc@{}}
\toprule
Method & Accuracy (\%) \\
\midrule
Full KV (0\% eviction) & 33.40 \\
EA & 31.41 \\
KeyDiff & 26.04 \\
TOVA & 30.22 \\
SnapKV & 30.62 \\
TriAttention & 31.21 \\
Q-Filters & 25.05 \\
Fast-KVzip & 32.01 \\
FAS-KV & 32.41 \\
FAS-KV + SSP & \textbf{34.19} \\
\bottomrule
\end{tabular}
\caption{LongBench-V2 accuracy on Qwen3-8B at 70\% KV-cache eviction.}
\label{tab:longbench_v2_comparison}
\end{table}

\begin{table}[t]
\centering
\footnotesize
\setlength{\tabcolsep}{3pt}
\begin{tabular}{@{}lrrrr@{}}
\toprule
& \multicolumn{2}{c}{Qwen3-8B} & \multicolumn{2}{c}{Llama-3.1-8B} \\
\cmidrule(lr){2-3}\cmidrule(l){4-5}
Direction only & 70\% & 80\% & 70\% & 80\% \\
\midrule
FAS-KV    & \textbf{46.06} & \textbf{43.26} & \textbf{41.97} & \textbf{40.25} \\
Q-Filters & 37.62 & 32.31 & 37.94 & 33.69 \\
\bottomrule
\end{tabular}
\caption{Controlled LongBench comparison at 70\% and 80\% eviction. Both methods use $s_i=(k_i^{\mathrm{post}})^\top w$ with matched calibration data and Sink-4.}
\label{tab:longbench_matched_directions}
\end{table}

Appendix~\ref{app:aime_results} further evaluates repeated decoding-time compression on AIME24 and AIME25. 
FAS-KV + SSP matches EA on AIME24 and outperforms it on AIME25, improving over the core FAS-KV scorer by 4.5 and 6.7 points, respectively.

\subsection{Runtime Efficiency}
\label{sec:efficiency}

Across both GPUs and batch sizes, FAS-KV adds 9.4--26.8 ms to prefill. At 70\% eviction, the median time from question input to the first answer token falls by more than half in all four configurations (Table~\ref{tab:prefill_runtime}).
On Qwen3-1.7B, FAS-KV + SSP increases throughput and reduces peak memory in every setting where Full KV fits on the RTX 5070 Ti (Table~\ref{tab:speed}). With four 32K-token contexts per batch, Full KV runs out of memory, while FAS-KV + SSP completes generation at 53.6 tokens/s with 13.93 GiB of peak GPU memory.

\begin{table}[t]
  \centering
  \footnotesize
  \renewcommand{\arraystretch}{0.96}
  \setlength{\tabcolsep}{3pt}
  \begin{tabular}{@{}lrrrr@{}}
    \toprule
    & \multicolumn{2}{c}{\shortstack{Qwen3-8B\\H200}}
    & \multicolumn{2}{c}{\shortstack{Qwen3-1.7B\\RTX 5070 Ti}} \\
    \cmidrule(lr){2-3}\cmidrule(l){4-5}
    Batch size & 1 & 2 & 1 & 2 \\
    \midrule
    Full KV prefill (ms) & 1494 & 3076 & 1908 & 3797 \\
    \midrule
    \multicolumn{5}{@{}l}{\textit{Prefill overhead relative to Full KV (ms)}} \\
    Q-Filters    &  13.1 &  22.3 &   7.7 &  12.9 \\
    KeyDiff      &  24.7 &  48.8 &  27.0 &  49.8 \\
    SnapKV       &  52.6 &  97.1 &  54.1 & 101.3 \\
    TOVA         &  24.9 &  44.2 &  15.1 &  22.7 \\
    EA           & 209.2 & 519.4 & 214.2 & 462.4 \\
    FAS-KV (Ours)       & 15.5 & 26.8 &  9.4 & 17.8 \\
    FAS-KV + SSP (Ours) & 16.5 & 32.2 & 10.3 & 19.8 \\
    \midrule
    \multicolumn{5}{@{}l}{\textit{Question processing + first answer token (ms)}} \\
    Full KV       & 87.6 & 117.5 & 83.4 & 110.1 \\
    70\% eviction & 38.1 &  50.0 & 33.0 &  41.7 \\
    \bottomrule
  \end{tabular}
  \caption{Median latency (ms) over 50 runs per configuration with a 24K-token context. In compressed runs, we evict 70\% of the context tokens from the KV cache.}
  \label{tab:prefill_runtime}
\end{table}

\begin{table}[t]
\centering
\small
\setlength{\tabcolsep}{3pt}
\begin{tabular}{rr rr rr}
\toprule
 & & \multicolumn{2}{c}{Throughput (tok/s)} & \multicolumn{2}{c}{Peak mem.\ (GiB)} \\
\cmidrule(lr){3-4}\cmidrule(lr){5-6}
$B$ & Ctx. & Full & \shortstack{FAS-KV\\+ SSP} & Full & \shortstack{FAS-KV\\+ SSP} \\
\midrule
1 & 8K  & 66.6  & 79.8 \, (1.20$\times$)  & 4.50  & 3.89 \, ($-$14\%) \\
1 & 16K & 46.2  & 62.5 \, (1.35$\times$)  & 5.78  & 4.56 \, ($-$21\%) \\
1 & 32K & 26.8  & 39.1 \, (1.46$\times$)  & 8.35  & 5.91 \, ($-$29\%) \\
\midrule
4 & 8K  & 119.4 & 180.5 \, (1.51$\times$) & 8.34  & 5.89 \, ($-$29\%) \\
4 & 16K & 66.8  & 108.8 \, (1.63$\times$) & 13.47 & 8.57 \, ($-$36\%) \\
4 & 32K & OOM   & \multicolumn{1}{l}{53.6 \, (--$\times$)} & OOM   & \multicolumn{1}{l}{13.93 \, (--\%)} \\
\bottomrule
\end{tabular}
\caption{Full KV vs.\ FAS-KV + SSP at 70\% eviction on Qwen3-1.7B (RTX 5070 Ti, 16\,GB). Each batch repeats one NarrativeQA context. Results average 10 runs generating 256 tokens. Parentheses show changes relative to Full KV. OOM denotes out of memory.}
\label{tab:speed}
\end{table}

\subsection{Ablation Study}

\paragraph{Held-out shared-direction validation.}
Figure~\ref{fig:shared_direction_heldout} evaluates whether the calibrated
direction transfers to actual LongBench generation without refitting. Across
the four models, FAS-KV achieves mean oracle-direction cosine similarities of
0.682--0.794 and realized-attention rank correlations of 0.685--0.733,
consistently outperforming both the uncanonicalized and query-based variants.
Prediction quality varies across both heads and tasks.
For example, in Qwen3-1.7B, the same KV head (layer~5, head~3) reaches
$\rho=0.87$ on QMSum but only $0.06$ on TriviaQA
(Appendix~\ref{app:prediction_errors}, Figure~\ref{fig:prediction_errors}).

\paragraph{Score-weight sensitivity.}
\leavevmode Table~\ref{tab:gamma} shows how LongBench scores vary with $\gamma$,
the weight of the shared-direction score relative to the log value norm in Eq.~\eqref{eq:value_aware_score}.
For FAS-KV on Qwen3-1.7B, $\gamma=0.1$ achieves the highest scores at
40--80\% eviction, whereas value-only scoring performs best at 20\%.
Combining both terms is particularly beneficial under aggressive eviction.

\begingroup
\setlength{\intextsep}{10pt plus 2pt minus 2pt}
\begin{table}[htbp]
\centering
\small
\setlength{\tabcolsep}{4pt}
\begin{tabular}{lcccc}
\toprule
Eviction & Value only & $\gamma=0.1$ & $\gamma=1$ & Direction only \\
\midrule
20\% & \textbf{37.72} & 36.93 & 36.90 & 36.77 \\
40\% & 37.17 & \textbf{37.46} & 37.40 & 37.29 \\
60\% & 34.74 & \textbf{37.40} & 36.78 & 36.76 \\
70\% & 32.09 & \textbf{36.65} & 36.17 & 35.16 \\
80\% & 27.38 & \textbf{34.80} & 34.15 & 32.62 \\
\bottomrule
\end{tabular}
\caption{FAS-KV score-weight sensitivity on LongBench with Qwen3-1.7B. Value-only and direction-only scores correspond to $\gamma=0$ and $\gamma\to\infty$, respectively.}
\label{tab:gamma}
\end{table}

\paragraph{Component ablations.}
Table~\ref{tab:component_ablation} reports component ablations on Qwen3-8B at 70\% compression. Adding SSP to FAS-KV improves RULER by 3.61 points. Canonicalization improves both benchmarks, while the fitted scale provides an additional gain, particularly on RULER. Removing the value norm causes the largest degradation, reducing the RULER score to 58.22.

\begin{table}[htbp]
  \centering
  \small
  \setlength{\tabcolsep}{3pt}
  \resizebox{\columnwidth}{!}{%
  \begin{tabular}{@{}lcc@{}}
    \toprule
    Method & LongBench & RULER 4K \\
    \midrule
    FAS-KV
    & 47.59 & 77.95 \\
    FAS-KV + SSP
    & 47.62 & 81.56 \\
    FAS-KV + all-layer SSP
    & 46.95 & 80.40 \\
    FAS-KV + alternative-layer SSP
    & 47.29 & 78.89 \\
    \midrule
    \quad w/o canonicalization
    & 47.02 & 77.10 \\
    \quad w/o fitted scale
    & 47.55 & 76.64 \\
    \quad w/o value norm
    & 46.06 & 58.22 \\
    Value norm only
    & 46.06 & 75.96 \\
    \bottomrule
  \end{tabular}%
  }
  \caption{Component ablations on Qwen3-8B at 70\% compression. Scores are
  mean LongBench and RULER-4K performance.}
  \label{tab:component_ablation}
\end{table}
\endgroup

The relative contributions of value magnitude and the shared
direction vary by task. On Qwen3-8B at 70\% eviction without SSP, the
value-only, direction-only, and combined scores are 71.2, 0.4, and 73.2
on \texttt{single\_3}, and 82.0, 92.6, and 95.1 on CWE, respectively.
Value-only scoring is stronger on \texttt{single\_3}, while direction-only
scoring is stronger on CWE. Combining them gives the highest score on both
tasks and improves the LongBench and RULER-4K averages
(Table~\ref{tab:component_ablation}).

\paragraph{Calibration robustness.}
Appendix~\ref{app:calibration_data} shows that FAS-KV remains effective
with limited calibration data and transfers across calibration domains and
languages.

\section{Related Work}
 
\paragraph{KV cache compression.}
KV-cache compression uses quantization~\citep{liu2024kivi,hooper2024kvquant},
architectural reduction~\citep{ainslie2023gqa,deepseekai2024deepseekv2}, or
channel pruning~\citep{xu2025think}. FAS-KV performs complementary \emph{token
eviction}.
 
\paragraph{Attention-score-based eviction.}
H2O, Scissorhands, TOVA, and SnapKV use past or recent attention
~\citep{zhang2023h2o,liu2023scissorhands,oren2024tova,li2024snapkv}, while
PyramidKV and AdaKV allocate budgets across layers or heads
~\citep{cai2024pyramidkv,feng2025adakv}. Requiring attention weights, and
sometimes the question before compression, limits fused-kernel compatibility
and query-agnostic eviction~\citep{dao2022flashattention,
dao2023flashattention2}.

\paragraph{FlashAttention compatibility.}
StreamingLLM retains sink and recent tokens; $L_2$-norm eviction, KeyDiff, and
ManifoldKV use attention-free geometry~\citep{xiao2024streamingllm,
devoto2024l2norm,park2025keydiff,datta2026manifoldkv}. EA and GVote model
future-query distributions with per-request computation
~\citep{devoto2025expectedattention,tang2025gvote}, while TriAttention uses
calibrated pre-RoPE query statistics and RoPE structure
~\citep{mao2026triattention}. FAS-KV instead learns one cumulative
future-attention direction per layer and per KV head offline and applies one key projection.

\paragraph{Learned eviction.}
Locret learns retaining heads~\citep{huang2024locret};
TrimKV/DBTrimKV study learned token retention
~\citep{bui2026dbtrimkv,bui2026cachewhatlasts}; KVP learns an eviction policy
~\citep{moschella2026kvp}; and Fast-KVzip uses auxiliary gates
~\citep{kim2026fastkvzip}. In contrast, FAS-KV requires only one
offline-calibrated direction per layer and KV head, without an auxiliary
prediction network.

\section{Conclusion}
We identified a shared future-attention direction in key space that becomes
transferable across inputs after RoPE boundary alignment. Based on this finding,
we introduced \textsc{FAS-KV}, which combines one projection per cached key with the log value norm to rank tokens for eviction. Across models and long-context benchmarks, it preserves
strong performance under aggressive compression with low eviction overhead.

\section{Limitations}
\textsc{FAS-KV} requires offline calibration with oracle future attention and must be
recalibrated for each model checkpoint. Its single shared direction captures
the dominant cross-input component of future-attention geometry, but cannot
represent input-specific residuals or adapt to input-dependent variation at
runtime. Although the calibrated direction shows transfer across calibration
domains and languages, downstream cross-lingual evaluation and generalization
beyond RoPE-based positional encodings remain untested. SSP encourages joint
subword retention but does not guarantee preservation of complete spans. SSP also depends on tokenizer-defined boundaries and a layer set selected from one synthetic sample; sensitivity to code, non-standard syntax, and unstructured inputs has not been evaluated. The base FAS-KV scorer does not rely on spans or this layer-selection step. Repeated compression is tested only in the limited AIME setting in Appendix~\ref{app:aime_results}, not in general streaming or multi-turn workloads.

\section*{Acknowledgments}
This work was supported by ABCI 3.0, which is provided by AIST and AIST Solutions.


\bibliography{custom}

\begin{thebibliography}{41}
\providecommand{\natexlab}[1]{#1}

\bibitem[{Ainslie et~al.(2023)Ainslie, Lee-Thorp, de~Jong, Zemlyanskiy, Lebr{\'o}n, and Sanghai}]{ainslie2023gqa}
Joshua Ainslie, James Lee-Thorp, Michiel de~Jong, Yury Zemlyanskiy, Federico Lebr{\'o}n, and Sumit Sanghai. 2023.
\newblock \href {https://aclanthology.org/2023.emnlp-main.298/} {{GQA}: Training generalized multi-query transformer models from multi-head checkpoints}.
\newblock In \emph{Proceedings of the 2023 Conference on Empirical Methods in Natural Language Processing}, pages 4895--4901.

\bibitem[{Bai et~al.(2024)Bai, Lv, Zhang, Lyu, Tang, Huang, Du, Liu, Zeng, Hou, Dong, Tang, and Li}]{bai2024longbench}
Yushi Bai, Xin Lv, Jiajie Zhang, Hongchang Lyu, Jiankai Tang, Zhidian Huang, Zhengxiao Du, Xiao Liu, Aohan Zeng, Lei Hou, Yuxiao Dong, Jie Tang, and Juanzi Li. 2024.
\newblock \href {https://doi.org/10.18653/v1/2024.acl-long.172} {{LongBench}: A bilingual, multitask benchmark for long context understanding}.
\newblock In \emph{Proceedings of the 62nd Annual Meeting of the Association for Computational Linguistics (Volume 1: Long Papers)}, pages 3119--3137.

\bibitem[{Bai et~al.(2025)Bai, Tu, Zhang, Peng, Wang, Lv, Cao, Xu, Hou, Dong, Tang, and Li}]{bai2025longbenchv2}
Yushi Bai, Shangqing Tu, Jiajie Zhang, Hao Peng, Xiaozhi Wang, Xin Lv, Shulin Cao, Jiazheng Xu, Lei Hou, Yuxiao Dong, Jie Tang, and Juanzi Li. 2025.
\newblock \href {https://doi.org/10.18653/v1/2025.acl-long.183} {{LongBench} v2: Towards deeper understanding and reasoning on realistic long-context multitasks}.
\newblock In \emph{Proceedings of the 63rd Annual Meeting of the Association for Computational Linguistics (Volume 1: Long Papers)}, pages 3639--3664.

\bibitem[{Bui et~al.(2026{\natexlab{a}})Bui, Nguyen, Cohan, and Ying}]{bui2026dbtrimkv}
Ngoc Bui, Hieu~Trung Nguyen, Arman Cohan, and Rex Ying. 2026{\natexlab{a}}.
\newblock \href {https://arxiv.org/abs/2605.09649} {Make each token count: Towards improving long-context performance with {KV} cache eviction}.
\newblock \emph{Preprint}, arXiv:2605.09649.

\bibitem[{Bui et~al.(2026{\natexlab{b}})Bui, Sharma, Lamba, Mishra, and Ying}]{bui2026cachewhatlasts}
Ngoc Bui, Shubham Sharma, Simran Lamba, Saumitra Mishra, and Rex Ying. 2026{\natexlab{b}}.
\newblock \href {https://proceedings.iclr.cc/paper_files/paper/2026/hash/c3887741ee4e655494770be95303383f-Abstract-Conference.html} {Cache what lasts: Token retention for memory-bounded {KV} cache in {LLMs}}.
\newblock In \emph{International Conference on Learning Representations}.

\bibitem[{Cai et~al.(2025)Cai, Zhang, Gao, Liu, Li, Liu, Lu, Xiong, Dong, Hu, and Xiao}]{cai2024pyramidkv}
Zefan Cai, Yichi Zhang, Bofei Gao, Yuliang Liu, Yucheng Li, Tianyu Liu, Keming Lu, Wayne Xiong, Yue Dong, Junjie Hu, and Wen Xiao. 2025.
\newblock \href {https://openreview.net/forum?id=ayi7qezU87} {{PyramidKV}: Dynamic {KV} cache compression based on pyramidal information funneling}.
\newblock In \emph{Second Conference on Language Modeling}.

\bibitem[{Dao(2024)}]{dao2023flashattention2}
Tri Dao. 2024.
\newblock \href {https://proceedings.iclr.cc/paper_files/paper/2024/hash/98ed250b203d1ac6b24bbcf263e3d4a7-Abstract-Conference.html} {{FlashAttention-2}: Faster attention with better parallelism and work partitioning}.
\newblock In \emph{International Conference on Learning Representations}.

\bibitem[{Dao et~al.(2022)Dao, Fu, Ermon, Rudra, and R{\'e}}]{dao2022flashattention}
Tri Dao, Daniel~Y. Fu, Stefano Ermon, Atri Rudra, and Christopher R{\'e}. 2022.
\newblock \href {https://proceedings.neurips.cc/paper/2022/hash/67d57c32e20fd0a7a302cb81d36e40d5-Abstract.html} {{FlashAttention}: Fast and memory-efficient exact attention with {IO}-awareness}.
\newblock In \emph{Advances in Neural Information Processing Systems}, volume~35, pages 16344--16359.

\bibitem[{Datta et~al.(2026)Datta, Neeraj, Paudel, Sharma, and Mukherjee}]{datta2026manifoldkv}
Debajyoti Datta, Trishala Neeraj, Bibek Paudel, Vyom Sharma, and Subhabrata Mukherjee. 2026.
\newblock \href {https://proceedings.mlr.press/v306/datta26b.html} {{ManifoldKV}: Training-free {KV} cache compression via {Euclidean} outlier detection}.
\newblock In \emph{Proceedings of the 43rd International Conference on Machine Learning}, volume 306 of \emph{Proceedings of Machine Learning Research}, pages 23070--23095.

\bibitem[{{DeepSeek-AI}(2024)}]{deepseekai2024deepseekv2}
{DeepSeek-AI}. 2024.
\newblock \href {https://arxiv.org/abs/2405.04434} {{DeepSeek-V2}: A strong, economical, and efficient mixture-of-experts language model}.
\newblock \emph{arXiv preprint arXiv:2405.04434}.

\bibitem[{Devoto et~al.(2025)Devoto, Jeblick, and J{\'e}gou}]{devoto2025expectedattention}
Alessio Devoto, Maximilian Jeblick, and Simon J{\'e}gou. 2025.
\newblock \href {https://arxiv.org/abs/2510.00636} {Expected attention: {KV} cache compression by estimating attention from future queries distribution}.
\newblock \emph{arXiv preprint arXiv:2510.00636}.

\bibitem[{Devoto et~al.(2024)Devoto, Zhao, Scardapane, and Minervini}]{devoto2024l2norm}
Alessio Devoto, Yu~Zhao, Simone Scardapane, and Pasquale Minervini. 2024.
\newblock \href {https://doi.org/10.18653/v1/2024.emnlp-main.1027} {A simple and effective {$L_2$} norm-based strategy for {KV} cache compression}.
\newblock In \emph{Proceedings of the 2024 Conference on Empirical Methods in Natural Language Processing}, pages 18476--18499. Association for Computational Linguistics.

\bibitem[{Feng et~al.(2025)Feng, Lv, Cao, Xie, and Zhou}]{feng2025adakv}
Yuan Feng, Junlin Lv, Yukun Cao, Xike Xie, and S.~Kevin Zhou. 2025.
\newblock \href {https://proceedings.neurips.cc/paper_files/paper/2025/hash/a40ff56daab9f4808b1e18350c8a11ce-Abstract-Conference.html} {{Ada-KV}: Optimizing {KV} cache eviction by adaptive budget allocation for efficient {LLM} inference}.
\newblock In \emph{Advances in Neural Information Processing Systems}, volume~38.

\bibitem[{{Gemma Team}(2025)}]{gemmateam2025gemma3}
{Gemma Team}. 2025.
\newblock \href {https://arxiv.org/abs/2503.19786} {{Gemma 3} technical report}.
\newblock \emph{Preprint}, arXiv:2503.19786.

\bibitem[{Godey et~al.(2025)Godey, Devoto, Zhao, Scardapane, Minervini, de~la Clergerie, and Sagot}]{godey2025qfilters}
Nathan Godey, Alessio Devoto, Yu~Zhao, Simone Scardapane, Pasquale Minervini, {\'E}ric de~la Clergerie, and Beno{\^i}t Sagot. 2025.
\newblock \href {https://arxiv.org/abs/2503.02812} {{Q-Filters}: Leveraging {QK} geometry for efficient {KV} cache compression}.
\newblock \emph{arXiv preprint arXiv:2503.02812}.

\bibitem[{Grattafiori et~al.(2024)}]{grattafiori2024llama3herdmodels}
Aaron Grattafiori et~al. 2024.
\newblock \href {https://arxiv.org/abs/2407.21783} {The {Llama 3} herd of models}.
\newblock \emph{Preprint}, arXiv:2407.21783.

\bibitem[{Hooper et~al.(2024)Hooper, Kim, Mohammadzadeh, Mahoney, Shao, Keutzer, and Gholami}]{hooper2024kvquant}
Coleman Hooper, Sehoon Kim, Hiva Mohammadzadeh, Michael~W. Mahoney, Yakun~Sophia Shao, Kurt Keutzer, and Amir Gholami. 2024.
\newblock \href {https://proceedings.neurips.cc/paper_files/paper/2024/hash/028fcbcf85435d39a40c4d61b42c99a4-Abstract-Conference.html} {{KVQuant}: Towards 10 million context length {LLM} inference with {KV} cache quantization}.
\newblock In \emph{Advances in Neural Information Processing Systems}, volume~37.

\bibitem[{Hsieh et~al.(2024)Hsieh, Sun, Kriman, Acharya, Rekesh, Jia, Zhang, and Ginsburg}]{hsieh2024ruler}
Cheng-Ping Hsieh, Simeng Sun, Samuel Kriman, Shantanu Acharya, Dima Rekesh, Fei Jia, Yang Zhang, and Boris Ginsburg. 2024.
\newblock \href {https://openreview.net/forum?id=kIoBbc76Sy} {{RULER}: What's the real context size of your long-context language models?}
\newblock In \emph{First Conference on Language Modeling}.

\bibitem[{Huang et~al.(2025)Huang, Yuan, Han, Xiao, and Liu}]{huang2024locret}
Yuxiang Huang, Binhang Yuan, Xu~Han, Chaojun Xiao, and Zhiyuan Liu. 2025.
\newblock \href {https://openreview.net/forum?id=YPVBCTBqHE} {{Locret}: Enhancing eviction in long-context {LLM} inference with trained retaining heads on consumer-grade devices}.
\newblock \emph{Transactions on Machine Learning Research}.

\bibitem[{Kamradt(2023)}]{kamradt2023niah}
Greg Kamradt. 2023.
\newblock \href {https://github.com/gkamradt/needle-in-a-haystack} {Needle in a haystack -- pressure testing {LLMs}}.
\newblock GitHub repository.

\bibitem[{Kim et~al.(2026)Kim, Han, and Yun}]{kim2026fastkvzip}
Jang-Hyun Kim, Dongyoon Han, and Sangdoo Yun. 2026.
\newblock \href {https://arxiv.org/abs/2601.17668} {{Fast KVzip}: Efficient and accurate {LLM} inference with gated {KV} eviction}.
\newblock \emph{Preprint}, arXiv:2601.17668.

\bibitem[{Li et~al.(2024)Li, Huang, Yang, Venkitesh, Locatelli, Ye, Cai, Lewis, and Chen}]{li2024snapkv}
Yuhong Li, Yingbing Huang, Bowen Yang, Bharat Venkitesh, Acyr Locatelli, Hanchen Ye, Tianle Cai, Patrick Lewis, and Deming Chen. 2024.
\newblock \href {https://proceedings.neurips.cc/paper_files/paper/2024/hash/28ab418242603e0f7323e54185d19bde-Abstract-Conference.html} {{SnapKV}: {LLM} knows what you are looking for before generation}.
\newblock In \emph{Advances in Neural Information Processing Systems}, volume~37.

\bibitem[{Liu et~al.(2023)Liu, Desai, Liao, Wang, Xie, Xu, Kyrillidis, and Shrivastava}]{liu2023scissorhands}
Zichang Liu, Aditya Desai, Fangshuo Liao, Weitao Wang, Victor Xie, Zhaozhuo Xu, Anastasios Kyrillidis, and Anshumali Shrivastava. 2023.
\newblock \href {https://proceedings.neurips.cc/paper_files/paper/2023/hash/a452a7c6c463e4ae8fbdc614c6e983e6-Abstract-Conference.html} {Scissorhands: Exploiting the persistence of importance hypothesis for {LLM} {KV} cache compression at test time}.
\newblock In \emph{Advances in Neural Information Processing Systems}, volume~36.

\bibitem[{Liu et~al.(2024)Liu, Yuan, Jin, Zhong, Xu, Braverman, Chen, and Hu}]{liu2024kivi}
Zirui Liu, Jiayi Yuan, Hongye Jin, Shaochen Zhong, Zhaozhuo Xu, Vladimir Braverman, Beidi Chen, and Xia Hu. 2024.
\newblock \href {https://proceedings.mlr.press/v235/liu24bz.html} {{KIVI}: A tuning-free asymmetric 2bit quantization for {KV} cache}.
\newblock In \emph{Proceedings of the 41st International Conference on Machine Learning}, volume 235 of \emph{Proceedings of Machine Learning Research}, pages 32332--32344.

\bibitem[{Mao et~al.(2026)Mao, Lin, Huang, Xie, Fu, Zhuang, Han, and Chen}]{mao2026triattention}
Weian Mao, Xi~Lin, Wei Huang, Yuxin Xie, Tianfu Fu, Bohan Zhuang, Song Han, and Yukang Chen. 2026.
\newblock \href {https://proceedings.mlr.press/v306/mao26e.html} {{TriAttention}: Efficient long reasoning with trigonometric {KV} compression}.
\newblock In \emph{Proceedings of the 43rd International Conference on Machine Learning}, volume 306 of \emph{Proceedings of Machine Learning Research}, pages 86458--86477.

\bibitem[{{Mathematical Association of America}(2024)}]{aime2024}
{Mathematical Association of America}. 2024.
\newblock American invitational mathematics examination ({AIME}) 2024 {I} and {II}.

\bibitem[{{Mathematical Association of America}(2025)}]{aime2025}
{Mathematical Association of America}. 2025.
\newblock American invitational mathematics examination ({AIME}) 2025 {I} and {II}.

\bibitem[{Merity et~al.(2017)Merity, Xiong, Bradbury, and Socher}]{merity2016pointer}
Stephen Merity, Caiming Xiong, James Bradbury, and Richard Socher. 2017.
\newblock \href {https://openreview.net/forum?id=Byj72udxe} {Pointer sentinel mixture models}.
\newblock In \emph{International Conference on Learning Representations}.

\bibitem[{{Meta AI}(2024)}]{meta2024llama32}
{Meta AI}. 2024.
\newblock \href {https://ai.meta.com/blog/llama-3-2-connect-2024-vision-edge-mobile-devices/} {{Llama 3.2}: Revolutionizing edge ai and vision with open, customizable models}.
\newblock Meta AI Blog.

\bibitem[{Moschella et~al.(2026)Moschella, Manduchi, and Sener}]{moschella2026kvp}
Luca Moschella, Laura Manduchi, and Ozan Sener. 2026.
\newblock \href {https://proceedings.mlr.press/v306/moschella26a.html} {Learning to evict from key-value cache}.
\newblock In \emph{Proceedings of the 43rd International Conference on Machine Learning}, volume 306 of \emph{Proceedings of Machine Learning Research}, pages 90470--90496.

\bibitem[{Oren et~al.(2024)Oren, Hassid, Yarden, Adi, and Schwartz}]{oren2024tova}
Matanel Oren, Michael Hassid, Nir Yarden, Yossi Adi, and Roy Schwartz. 2024.
\newblock \href {https://aclanthology.org/2024.emnlp-main.1043/} {Transformers are multi-state {RNNs}}.
\newblock In \emph{Proceedings of the 2024 Conference on Empirical Methods in Natural Language Processing}, pages 18724--18741. Association for Computational Linguistics.

\bibitem[{Park et~al.(2025)Park, Jones, Morse, Goel, Lee, and Lott}]{park2025keydiff}
Junyoung Park, Dalton Jones, Matthew Morse, Raghavv Goel, Mingu Lee, and Christopher Lott. 2025.
\newblock \href {https://proceedings.neurips.cc/paper_files/paper/2025/hash/0907335ecf28faf15be54485dbcbe70e-Abstract-Conference.html} {{KeyDiff}: Key similarity-based {KV} cache eviction for long-context {LLM} inference in resource-constrained environments}.
\newblock In \emph{Advances in Neural Information Processing Systems}, volume~38.

\bibitem[{Raffel et~al.(2020)Raffel, Shazeer, Roberts, Lee, Narang, Matena, Zhou, Li, and Liu}]{raffel2020t5}
Colin Raffel, Noam Shazeer, Adam Roberts, Katherine Lee, Sharan Narang, Michael Matena, Yanqi Zhou, Wei Li, and Peter~J. Liu. 2020.
\newblock \href {https://jmlr.org/papers/v21/20-074.html} {Exploring the limits of transfer learning with a unified text-to-text transformer}.
\newblock \emph{Journal of Machine Learning Research}, 21(140):1--67.

\bibitem[{See et~al.(2017)See, Liu, and Manning}]{see2017get}
Abigail See, Peter~J. Liu, and Christopher~D. Manning. 2017.
\newblock \href {https://doi.org/10.18653/v1/P17-1099} {Get to the point: Summarization with pointer-generator networks}.
\newblock In \emph{Proceedings of the 55th Annual Meeting of the Association for Computational Linguistics (Volume 1: Long Papers)}, pages 1073--1083.

\bibitem[{Su et~al.(2024)Su, Ahmed, Lu, Pan, Bo, and Liu}]{su2024roformer}
Jianlin Su, Murtadha Ahmed, Yu~Lu, Shengfeng Pan, Wen Bo, and Yunfeng Liu. 2024.
\newblock \href {https://doi.org/10.1016/j.neucom.2023.127063} {{RoFormer}: Enhanced transformer with rotary position embedding}.
\newblock \emph{Neurocomputing}, 568:127063.

\bibitem[{Tang et~al.(2025)Tang, Liu, Xu, and Huang}]{tang2025gvote}
Chenxia Tang, Jianchun Liu, Hongli Xu, and Liusheng Huang. 2025.
\newblock \href {https://arxiv.org/abs/2509.03136} {Adaptive {KV}-cache compression without manually setting budget}.
\newblock \emph{arXiv preprint arXiv:2509.03136}.

\bibitem[{{Wikimedia Foundation}(2023)}]{wikimedia2023wikipedia}
{Wikimedia Foundation}. 2023.
\newblock \href {https://huggingface.co/datasets/wikimedia/wikipedia} {Wikipedia dumps}.
\newblock Hugging Face dataset.
\newblock November 1, 2023 dump; English and Japanese configurations.

\bibitem[{Xiao et~al.(2024)Xiao, Tian, Chen, Han, and Lewis}]{xiao2024streamingllm}
Guangxuan Xiao, Yuandong Tian, Beidi Chen, Song Han, and Mike Lewis. 2024.
\newblock \href {https://proceedings.iclr.cc/paper_files/paper/2024/hash/5e5fd18f863cbe6d8ae392a93fd271c9-Abstract-Conference.html} {Efficient streaming language models with attention sinks}.
\newblock In \emph{International Conference on Learning Representations}.

\bibitem[{Xu et~al.(2025)Xu, Jie, Dong, Wang, Lu, Zhou, Saha, Xiong, and Sahoo}]{xu2025think}
Yuhui Xu, Zhanming Jie, Hanze Dong, Lei Wang, Xudong Lu, Aojun Zhou, Amrita Saha, Caiming Xiong, and Doyen Sahoo. 2025.
\newblock \href {https://proceedings.iclr.cc/paper_files/paper/2025/hash/8edb116d5b288b6a9bba4c16ab647702-Abstract-Conference.html} {{ThinK}: Thinner key cache by query-driven pruning}.
\newblock In \emph{International Conference on Learning Representations}.

\bibitem[{Yang et~al.(2025)Yang, Li, Yang, Zhang, Hui, Zheng, Yu, Gao, Huang, Lv, Zheng, Liu, Zhou, Huang, Hu, Ge, Wei, Lin, Tang, Yang, Tu, Zhang, Yang, Yang, Zhou, Zhou, Lin, Dang, Bao, Yang, Yu, Deng, Li, Xue, Li, Zhang, Wang, Zhu, Men, Gao, Liu, Luo, Li, Tang, Yin, Ren, Wang, Zhang, Ren, Fan, Su, Zhang, Zhang, Wan, Liu, Wang, Cui, Zhang, Zhou, and Qiu}]{yang2025qwen3technicalreport}
An~Yang, Anfeng Li, Baosong Yang, Beichen Zhang, Binyuan Hui, Bo~Zheng, Bowen Yu, Chang Gao, Chengen Huang, Chenxu Lv, Chujie Zheng, Dayiheng Liu, Fan Zhou, Fei Huang, Feng Hu, Hao Ge, Haoran Wei, Huan Lin, Jialong Tang, and 41 others. 2025.
\newblock \href {https://arxiv.org/abs/2505.09388} {{Qwen3} technical report}.
\newblock \emph{Preprint}, arXiv:2505.09388.

\bibitem[{Zhang et~al.(2023)Zhang, Sheng, Zhou, Chen, Zheng, Cai, Song, Tian, R{\'e}, Barrett, Wang, and Chen}]{zhang2023h2o}
Zhenyu Zhang, Ying Sheng, Tianyi Zhou, Tianlong Chen, Lianmin Zheng, Ruisi Cai, Zhao Song, Yuandong Tian, Christopher R{\'e}, Clark Barrett, Zhangyang Wang, and Beidi Chen. 2023.
\newblock \href {https://proceedings.neurips.cc/paper_files/paper/2023/hash/6ceefa7b15572587b78ecfcebb2827f8-Abstract.html} {{$H_2O$}: Heavy-hitter oracle for efficient generative inference of large language models}.
\newblock In \emph{Advances in Neural Information Processing Systems}, volume~36, pages 34661--34710.

\end{thebibliography}

\appendix
\section{Exact Linearity of Centered Log-Attention}
\label{app:exact_linearity}

For completeness, we derive Eq.~\eqref{eq:exact_log_attention}.
For a fixed future query $q_t$, define the softmax normalizer
\begin{equation}
Z_t
=
\sum_{j\leq t}
\exp\left(
q_t^\top k_j/\sqrt{d}
\right).
\end{equation}

The log-attention assigned to token $i$ is
\begin{equation}
\log a_{t,i}
=
\frac{q_t^\top k_i}{\sqrt{d}}
-
\log Z_t.
\label{eq:appendix_log_attention}
\end{equation}

Averaging over cached tokens gives
\begin{equation}
\frac{1}{N}
\sum_{i=1}^{N}
\log a_{t,i}
=
\frac{q_t^\top\bar{k}}{\sqrt{d}}
-
\log Z_t.
\label{eq:appendix_mean_log_attention}
\end{equation}

Subtracting Eq.~\eqref{eq:appendix_mean_log_attention} from
Eq.~\eqref{eq:appendix_log_attention} yields
\begin{equation}
\widetilde{a}_{t,i}
=
(k_i-\bar{k})^\top
\frac{q_t}{\sqrt{d}},
\end{equation}

which proves the exact linearity of centered log-attention.
The approximation in Eq.~\eqref{eq:cumulative_linear_model} is required only
because cumulative future importance aggregates several independently
normalized attention distributions.

\section{Additional Theoretical Analyses}

\subsection{Relationship Between Direct Contribution and Deletion Effect}
\label{app:deletion_effect}

Our oracle importance measures the average magnitude of a token's direct
contribution to the attention output, rather than the change caused by
deleting that token. We clarify the relationship between these two quantities.

Consider a fixed future query at position $t$. The attention output before
the output projection is
\begin{equation}
    o_t = \sum_j a_{t,j}v_j.
\end{equation}
For a cached token $i$, define the renormalized weighted average of the
remaining values as
\begin{equation}
    \bar v_{t,-i}
    = \sum_{j\ne i}
      \frac{a_{t,j}}{1-a_{t,i}}v_j,
\end{equation}
where $a_{t,i}<1$ is assumed. Since
\begin{equation}
    o_t = a_{t,i}v_i
          +(1-a_{t,i})\bar v_{t,-i},
\end{equation}
deleting token $i$ and renormalizing the remaining attention weights gives
\begin{equation}
    o_t^{(-i)} = \bar v_{t,-i}.
\end{equation}
Consequently, the exact output change is
\begin{align}
    \Delta_{t,i}
    &= \|o_t-o_t^{(-i)}\|_2 \\
    &= a_{t,i}\|v_i-\bar v_{t,-i}\|_2.
\end{align}

In contrast, the direct-contribution magnitude used by our oracle is
\begin{equation}
    d_{t,i}=a_{t,i}\|v_i\|_2.
\end{equation}
By the reverse triangle inequality,
\begin{equation}
    |\Delta_{t,i}-d_{t,i}|
    \le a_{t,i}\|\bar v_{t,-i}\|_2.
\end{equation}

For a future window $\mathcal{F}$ of length $H=|\mathcal{F}|$, define the
average deletion effect and direct contribution as
\begin{equation}
    \widetilde D_i
    =\frac{1}{H}\sum_{t\in\mathcal{F}}\Delta_{t,i},
    \qquad
    D_i
    =\frac{1}{H}\sum_{t\in\mathcal{F}}d_{t,i}.
\end{equation}
Then
\begin{equation}
    |\widetilde D_i-D_i|
    \le \frac{1}{H}
    \sum_{t\in\mathcal{F}}
    a_{t,i}\|\bar v_{t,-i}\|_2.
\end{equation}

This bound requires no isotropy assumption on the value vectors. It shows that
the discrepancy depends on the weighted mean of the remaining values, which
can reflect both their directions and their cancellation. When this mean is
small, direct contribution closely approximates the deletion effect.
However, the bound does not guarantee that the two quantities induce the same
token ranking in general.

Thus, $D_i$ is an exact measure of average direct-contribution magnitude, but
not an exact measure of deletion impact. This analysis concerns a single
attention head before the output projection and does not characterize the full
downstream effect of cache eviction.

\subsection{Local Linearization of Cumulative Future Attention}
\label{app:local_linearization}

We provide a mathematical explanation for the approximately linear
relationship between centered log cumulative attention and cached keys. For a
fixed input, let $\{k_i\}_{i=1}^{N}$ be the cached keys and let $\mathcal{F}$
be the future observation window, with $H=|\mathcal{F}|$. Define
\begin{equation}
    b_t=\frac{q_t}{\sqrt d},
    \qquad
    c_t=\frac{1}{H Z_t},
\end{equation}
where $Z_t$ is the softmax normalizer for the future query at position $t$.
Holding the observed queries and normalizers fixed, the average attention
received by cached token $i$ can be written as
\begin{equation}
    \alpha_i
    =\sum_{t\in\mathcal{F}}c_t\exp(b_t^\top k_i).
\end{equation}

To account for the numerical offset $\delta$ in the regression target, define
\begin{equation}
    f(k)=
    \log\left(
      \delta+\sum_{t\in\mathcal{F}}
      c_t\exp(b_t^\top k)
    \right).
\end{equation}
Thus, the centered log cumulative attention is
\begin{equation}
    y_i=f(k_i)-\frac1N\sum_{j=1}^N f(k_j).
\end{equation}

For convenience, introduce a dummy term with $b_0=0$ and $c_0=\delta$, and
let $\mathcal{F}_0=\mathcal{F}\cup\{0\}$. Define
\begin{equation}
    \pi_t(k)=
    \frac{c_t\exp(b_t^\top k)}
    {\sum_{s\in\mathcal{F}_0}c_s\exp(b_s^\top k)}.
\end{equation}
The gradient of $f$ is then
\begin{equation}
    \nabla f(k)=
    \sum_{t\in\mathcal{F}_0}\pi_t(k)b_t.
\end{equation}

Its Hessian is
\begin{align}
    \nabla^2 f(k)
    &=\sum_{t\in\mathcal{F}_0}\pi_t(k)b_tb_t^\top \\
    &\quad-\nabla f(k)\nabla f(k)^\top \\
    &=\operatorname{Cov}_{\pi(k)}(b_t).
\end{align}
Therefore, the curvature of log cumulative attention is governed by the
weighted covariance of the future-query directions, rather than requiring the
queries to be collinear.

Let
\begin{equation}
    \bar k=\frac1N\sum_{i=1}^N k_i,
    \qquad
    w_{\mathrm{loc}}=\nabla f(\bar k).
\end{equation}
A second-order Taylor expansion around $\bar k$ gives
\begin{equation}
    f(k_i)=f(\bar k)
    +(k_i-\bar k)^\top w_{\mathrm{loc}}
    +r_i.
\end{equation}

If
\begin{equation}
    \|\nabla^2 f(k)\|_{\mathrm{op}}\le M
\end{equation}
throughout the convex hull of the cached keys, the remainder satisfies
\begin{equation}
    |r_i|\le
    \frac M2\|k_i-\bar k\|_2^2.
\end{equation}

Centering across cached tokens eliminates the constant term and yields
\begin{equation}
    y_i=(k_i-\bar k)^\top w_{\mathrm{loc}}
        +(r_i-\bar r),
\end{equation}
where
\begin{equation}
    \bar r=\frac1N\sum_{j=1}^N r_j.
\end{equation}
Consequently,
\begin{align}
    |r_i-\bar r|
    \le\frac M2\left(
       \|k_i-\bar k\|_2^2
       +\frac1N\sum_{j=1}^N
        \|k_j-\bar k\|_2^2
    \right).
\end{align}

This establishes a local linear approximation whose error is controlled by
the curvature along the observed key displacements. In particular, small
variation among future queries in the relevant key-space directions leads to
small curvature and a more accurate linear approximation.

The locally derived direction $w_{\mathrm{loc}}$ provides a mathematical
explanation for linear predictability, whereas the regularized regression
direction $w^\star$ used in our method is fitted directly to the observed
targets. The strong empirical predictability reported in the main paper does
not require the above error bound to be uniformly small.

Finally, this analysis concerns linearity within a fixed input. The existence
of a dominant direction shared across inputs after RoPE boundary
canonicalization remains an empirical observation, rather than a consequence
of the Taylor expansion.

\section{Grouped-Query Attention}
\label{app:gqa}

In grouped-query attention, multiple query heads share one KV head.
Let $\mathcal{G}(h)$ denote the set of query heads associated with KV head
$h$.
For calibration sample $m$, let $H_m=|\mathcal{F}_m|$. We aggregate attention
over both future positions and the corresponding query heads by averaging over
both axes:
\begin{equation}
\alpha_{m,i}^{(h)}
=
\frac{1}{|\mathcal{G}(h)|H_m}
\sum_{g\in\mathcal{G}(h)}
\sum_{t\in\mathcal{F}_m}
a_{m,t,i}^{(g)}.
\label{eq:gqa_future_importance}
\end{equation}

The centered log target is then constructed from
$\alpha_{m,i}^{(h)}$, and one shared canonical direction is estimated for each
KV head.
The resulting score represents the combined future demand of all
query heads that access the same cached keys and values.

\section{Implementation, Hardware, and Compression Ratios}
\label{app:hardware_compression}

\paragraph{Implementation.}
We implemented \textsc{FAS-KV} by extending the open-source KVPress framework
(v0.5.3, commit \texttt{fa7a0dc1d6d5332a8ab00d03c67de7182c6846b0})
\citep{devoto2025expectedattention}. We integrated our scoring and SSP components into its press interface. Wherever supported by the
baseline implementation, we used the same evaluation pipeline, with matched
models, datasets, decoding configurations, and compression settings. The
original Q-Filters implementation did not support Qwen3, so we extended it to the Qwen3 architecture and fitted its projection using the same calibration data.
We evaluated Fast-KVzip on Qwen3-8B
using the authors' code and released gates~\citep{kim2026fastkvzip}.
For TriAttention, we reimplemented the scoring rule in Eqs.~(6--13) of
\citet{mao2026triattention} and estimated its query statistics on the same
calibration corpus as FAS-KV. We applied the scorer once after prefilling the
context, before appending the question.
In the main experiments, each baseline uses its original sink-token policy.
Following EA, FAS-KV always retains the first four tokens.
Table~\ref{tab:longbench_matched_directions} compares the FAS-KV and Q-Filters directions with the same four tokens protected and the same per-head cache budget.

\paragraph{Hardware and compression ratios.}
We run all experiments with the 8B models on an NVIDIA H200 GPU and those with
the smaller models on an NVIDIA GeForce RTX 5070 Ti GPU.
The compression ratio $r$ is the fraction of context tokens
evicted from each KV head's cache after prefilling. For a context of $T$ tokens,
each head retains $K=\lfloor(1-r)T\rfloor$ tokens, or approximately 30\% of the
context when $r=0.7$.

\section{Calibration Data and Continuation Construction}
\label{app:calibration_data}

We use ordinary continuous English documents rather than question--answer
data. We sample 64 documents from each of the three sources, giving
$M=64\times3=192$ calibration samples in total. The sources are listed in
Table~\ref{tab:calibration_corpora}.

\begin{table*}[t]
\centering
\scriptsize
\setlength{\tabcolsep}{5pt}
\begin{tabular}{llll}
\toprule
\textbf{Source} & \textbf{Hugging Face dataset / configuration} &
\textbf{Field} & \textbf{Content} \\
\midrule
\texttt{en\_wikipedia}~\citep{wikimedia2023wikipedia} & \texttt{wikimedia/wikipedia} / \texttt{20231101.en}
& \texttt{text} & English Wikipedia articles \\
\texttt{en\_c4}~\citep{raffel2020t5} & \texttt{allenai/c4} / \texttt{en}
& \texttt{text} & General web documents \\
\texttt{en\_cnn\_dailymail}~\citep{see2017get} & \texttt{abisee/cnn\_dailymail} / \texttt{3.0.0}
& \texttt{article} & News articles \\
\bottomrule
\end{tabular}
\caption{Calibration corpora and fields used to obtain continuous text.}
\label{tab:calibration_corpora}
\end{table*}

Each document is tokenized and split into a cached context followed by its
natural continuation. For a sufficiently long document, we use
\begin{equation}
\underbrace{[z_1,\ldots,z_{4096}]}_{\text{cached context}}
\quad+
\underbrace{[z_{4097},\ldots,z_{4224}]}_{\text{pseudo-future}}.
\end{equation}

Thus, the context contains 4,096 tokens and the future teacher contains 128
tokens. If the original document exceeds 4,224 tokens, only its first 4,224
tokens are used; we do not sample a random window or take a document-final
window. For shorter documents, we set
\begin{equation}
H=\min\left(128,L-T_{\min}\right),
\end{equation}

where $T_{\min}=128$ is the required minimum context length, and use the final
$H$ tokens as the pseudo-future. Documents of at least 256 tokens normally retain
the full 128-token future window. This produces continuation-style calibration
from natural prose, rather than the LongBench format in which a question is
placed after a long context.

More generally, let calibration sample $m$ have total length $L_m$, and let
$T_m$ be the position of its first pseudo-future token. Using zero-based
positions, the cached context occupies $0,\ldots,T_m-1$, and we define
\begin{equation}
\begin{aligned}
H_m
&=\min(128,L_m-T_m),
\\
\mathcal{F}_m
&=\{T_m,\ldots,T_m+H_m-1\}.
\end{aligned}
\label{eq:calibration_future_window}
\end{equation}

\subsection{Calibration-Set Size Ablation}
\label{app:calibration_size_ablation}

Table~\ref{tab:calibration_size} reports the sensitivity of FAS-KV to the
total number of calibration samples across the three sources.

\begin{table}[t]
\centering
\small
\setlength{\tabcolsep}{3pt}
\begin{tabular}{@{}lrrrrr@{}}
\toprule
\shortstack[l]{Calibration\\samples (total)}
& \multicolumn{5}{c}{Compression ratio} \\
\cmidrule(l){2-6}
& 0.2 & 0.4 & 0.6 & 0.7 & 0.8 \\
\midrule
24  & 84.47 & 74.95 & 54.01 & 44.32 & 36.77 \\
96  & 84.56 & 75.42 & 53.30 & 43.44 & 36.04 \\
192 & 84.83 & 75.11 & 53.73 & 42.90 & 35.85 \\
768 & 84.44 & 75.36 & 54.82 & 43.91 & 35.89 \\
\midrule
Range
& 0.39 & 0.47 & 1.52 & 1.42 & 0.92 \\
$\Delta_{24\rightarrow768}$
& $-0.03$ & $+0.41$ & $+0.81$ & $-0.41$ & $-0.88$ \\
\bottomrule
\end{tabular}
\caption{
Effect of the total number of calibration samples on RULER-4K performance
for Qwen3-1.7B without SSP.
Scores are averaged over all 13 RULER subtasks.
}
\label{tab:calibration_size}
\end{table}

\subsection{Calibration-Domain Transfer}
\label{app:calibration_domain_transfer}

Table~\ref{tab:calibration_domain_transfer} evaluates whether the shared
direction depends on the language, domain, or size of the calibration set.
Using 192 Japanese Wikipedia samples slightly improves the average LongBench
score over the default English-domain mixture, despite the language mismatch.
Even a calibration set of only four Japanese Wikipedia samples remains within
0.52 points of the default setting on average, indicating that FAS-KV requires
little calibration data and transfers across calibration domains and
languages.

\begin{table}[t]
\centering
\scriptsize
\setlength{\tabcolsep}{2pt}
\resizebox{\columnwidth}{!}{%
\begin{tabular}{@{}lrrrrrr@{}}
\toprule
Calibration data (samples) & 20\% & 40\% & 60\% & 70\% & 80\% & Mean \\
\midrule
Japanese Wikipedia (4)
& 36.64 & 37.05 & 36.87 & 35.91 & 34.17 & 36.13 \\
Japanese Wikipedia (192)
& 36.88 & 37.29 & 37.82 & 36.74 & 35.28 & 36.80 \\
English C4 only (192)
& 37.09 & 37.53 & 37.14 & 36.41 & 34.27 & 36.49 \\
Default mixture (192)
& 36.93 & 37.46 & 37.40 & 36.65 & 34.80 & 36.65 \\
\bottomrule
\end{tabular}%
}
\caption{Effect of calibration language, domain, and sample count on mean
LongBench performance for Qwen3-1.7B without SSP. The default mixture is the
192-sample calibration set used in the main experiments.}
\label{tab:calibration_domain_transfer}
\end{table}

\subsection{Sensitivity to Future-Window Length}
\label{app:future_window}

We vary the number of future tokens $H$ used to estimate
$\alpha_i^{\mathrm{future}}$ during calibration
(Table~\ref{tab:future_window}). Across the five lengths tested, the
LongBench average varies by 0.31 points for Llama-3.2-1B-Instruct and
0.71 points for Qwen3-1.7B. The default $H=128$ is within 0.14 points
of the best score for each model.

\begin{table}[t]
\centering
\footnotesize
\setlength{\tabcolsep}{2pt}
\begin{tabular}{@{}lrrrrr@{}}
\toprule
Model & $H{=}32$ & $64$ & $128$ & $256$ & $512$ \\
\midrule
Llama-3.2-1B-Instruct & 23.22 & 22.91 & \underline{23.16} & 22.92 & 23.22 \\
Qwen3-1.7B            & 36.79 & 36.59 & \underline{36.65} & 36.43 & 36.08 \\
\bottomrule
\end{tabular}
\caption{Sensitivity to the calibration window length $H$. Scores are
LongBench averages over 16 tasks at 70\% compression for FAS-KV.
Underlining marks the default window ($H=128$).}
\label{tab:future_window}
\end{table}

\section{Calibration Cost}
\label{app:calibration_cost}

For Qwen3-1.7B on an NVIDIA GeForce RTX 5070 Ti GPU, the offline FAS-KV
calibration takes approximately 4 minutes. The offline search for the
SSP layer set takes approximately 10 minutes. Both procedures are performed
only once per model checkpoint and introduce no additional per-request
calibration cost at inference time.

\section{Oracle-Direction Fitting}
\label{app:oracle_fitting}

We fit an oracle direction independently for every calibration sample, model
layer, and KV head. Future attention is aggregated over the query heads sharing
the KV head as in Eq.~\eqref{eq:gqa_future_importance}. The first
$K_{\mathrm{sink}}=4$ sink tokens are excluded because they are permanently
retained. Suppressing layer and KV-head indices, the fitted token set is
\begin{equation}
\mathcal{I}_m
=
\{K_{\mathrm{sink}},\ldots,T_m-1\},
\qquad
N_m=|\mathcal{I}_m|.
\end{equation}

The centered log-attention target is
\begin{equation}
\begin{aligned}
y_{m,i}
&=
\log(\alpha_{m,i}+\delta)
-
\frac{1}{N_m}
\sum_{j\in\mathcal{I}_m}
\log(\alpha_{m,j}+\delta),
\\[-2pt]
&\hspace{8em} i\in\mathcal{I}_m,
\end{aligned}
\label{eq:calibration_target}
\end{equation}

where $\delta=10^{-6}$. We center the post-RoPE keys over the same set:
\begin{equation}
x_{m,i}
=
k_{m,i}^{\mathrm{post}}
-
\frac{1}{N_m}
\sum_{j\in\mathcal{I}_m}k_{m,j}^{\mathrm{post}},
\qquad i\in\mathcal{I}_m.
\label{eq:oracle_design_matrix}
\end{equation}

Writing $\mathcal{I}_m=\{i_1,\ldots,i_{N_m}\}$, define
\begin{equation}
X_m
=
\begin{bmatrix}
x_{m,i_1}^{\top}\\
\vdots\\
x_{m,i_{N_m}}^{\top}
\end{bmatrix}
\in\mathbb{R}^{N_m\times d}.
\label{eq:oracle_design_matrix_stack}
\end{equation}

We do not normalize individual keys. If $\{\nu_{m,j}\}$ are the eigenvalues
of $X_m^\top X_m$, the covariance-scaled ridge coefficient is
\begin{equation}
\lambda_m
=
0.1\,
\operatorname{median}
\{\nu_{m,j}:\nu_{m,j}>0\}.
\label{eq:ridge_regularization}
\end{equation}

The oracle direction and its unit-normalized form are
\begin{subequations}
\begin{align}
w_m^{\star,\mathrm{post}}
&=
\left(X_m^\top X_m+\lambda_m I\right)^{-1}X_m^\top y_m,
\label{eq:ridge_direction}
\\
\widehat{w}_m^{\mathrm{post}}
&=
\frac{w_m^{\star,\mathrm{post}}}
{\lVert w_m^{\star,\mathrm{post}}\rVert_2}.
\label{eq:normalized_oracle_direction}
\end{align}
\end{subequations}

The normalization separates the predictive orientation from
sample-dependent variation in the fitted norm.

\section{Boundary Canonicalization and Scale-Fit Details}

\subsection{Boundary Canonicalization}
\label{app:boundary_canonicalization}

Let $R(t)$ denote the RoPE rotation at position $t$, such that
$k_i^{\mathrm{post}}=R(i)k_i^{\mathrm{pre}}$ and
$q_t^{\mathrm{post}}=R(t)q_t^{\mathrm{pre}}$. Applying the common boundary
rotation $R(-T)$ gives
\begin{subequations}
\begin{align}
k_{i\mid T}
&=
R(-T)k_i^{\mathrm{post}}
=R(i-T)k_i^{\mathrm{pre}},
\\
q_{t\mid T}
&=
R(-T)q_t^{\mathrm{post}}
=R(t-T)q_t^{\mathrm{pre}}.
\end{align}
\end{subequations}

Since $R(-T)$ is orthogonal, this coordinate change preserves every
query--key inner product:
\begin{equation}
\left(k_i^{\mathrm{post}}\right)^\top q_t^{\mathrm{post}}
=
k_{i\mid T}^\top q_{t\mid T}.
\end{equation}

The same argument applies to the centered calibration keys. Define
$x_{m,i\mid T_m}=R(-T_m)x_{m,i}$ and
$u_m=R(-T_m)\widehat{w}_m^{\mathrm{post}}$. Then
\begin{equation}
x_{m,i}^\top\widehat{w}_m^{\mathrm{post}}
=
x_{m,i\mid T_m}^\top u_m.
\end{equation}

Thus, canonicalization preserves all oracle fitted scores while expressing
each sample using relative phases $i-T_m$. Rotating each key separately by
$R(-i)$ would instead place different tokens in different coordinate frames,
so one shared scoring direction could no longer reproduce their scores. RoPE
also preserves vector norms, and hence $\lVert u_m\rVert_2=1$.

\subsection{Sample-Wise Scalar Fit}
\label{app:beta_fitting}

Restoring layer and KV-head indices, transport the shared unit direction to
sample $m$ and form its one-dimensional prediction:
\begin{equation}
\mu_{\ell,h}^{(m)}
=
R(T_m)\mu_{\ell,h},
\qquad
z_{m,\ell,h}
=
X_{m,\ell,h}\mu_{\ell,h}^{(m)}.
\end{equation}

The sample-wise ridge fit and its closed form are
\begin{equation}
\begin{aligned}
\beta_{m,\ell,h}
&=
\arg\min_{\beta\in\mathbb{R}}
\left\lVert
y_{m,\ell,h}-\beta z_{m,\ell,h}
\right\rVert_2^2
+\lambda_\beta\beta^2,
\\
&=
\frac{
z_{m,\ell,h}^{\top}y_{m,\ell,h}
}{
z_{m,\ell,h}^{\top}z_{m,\ell,h}+\lambda_\beta
}.
\end{aligned}
\end{equation}

We use $\lambda_\beta=10^{-2}$ and store
\begin{equation}
\beta_{\ell,h}^{\mathrm{fit}}
=
\frac{1}{M}\sum_{m=1}^{M}\beta_{m,\ell,h},
\qquad
w_{\ell,h}^{\mathrm{FAS,can}}
=
\beta_{\ell,h}^{\mathrm{fit}}\mu_{\ell,h}.
\end{equation}

\section{Eviction Procedure}
\label{app:eviction}

Suppose the prefilled cache contains $T$ tokens and the compression
ratio is $r$. We set the total retained-cache budget to
\[
K = \lfloor (1-r)T \rfloor .
\]
The first $K_{\mathrm{sink}}=4$ tokens are always retained and are
included in this total budget. Let
\[
\mathcal{C}=\{K_{\mathrm{sink}},\ldots,T-1\}
\]
be the remaining candidate tokens. We select
\[
\mathcal{I}_{\mathrm{score}}
=
\operatorname{TopK}_{i\in\mathcal{C}}
\left(s_i, K-K_{\mathrm{sink}}\right),
\]
and retain
\[
\mathcal{I}_{\mathrm{keep}}
=
\mathcal{I}_{\mathrm{score}}
\cup
\{0,\ldots,K_{\mathrm{sink}}-1\}.
\]
Keys and values outside $\mathcal{I}_{\mathrm{keep}}$ are removed.
For SSP, the propagated score $\widetilde{s}_i$ is used in place of
$s_i$ at the selected layers.

\section{Selective Span Propagation}
\label{sec:layer_selective_span}

Token-wise KV eviction may retain only part of an identifier, number, or name
that the tokenizer splits into multiple subwords. We address this failure mode
with Selective Span Propagation (SSP). The runtime procedure neither
searches the input text nor applies named-entity recognition. Instead, it forms
lexical spans directly from tokenizer boundary markers and propagates the
largest token score within each span. This encourages related subwords to be
retained together, but does not force a span to be selected as an indivisible
unit.

\begin{figure}[t]
    \centering
    \includegraphics[
        width=\columnwidth
    ]{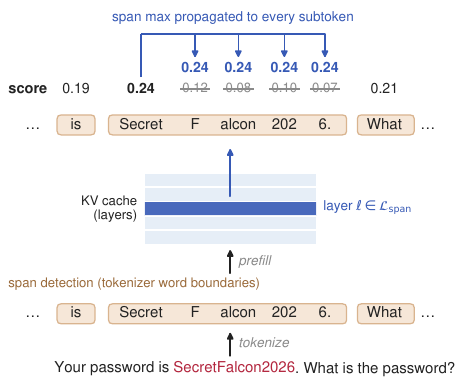}
    \caption{Overview of tokenizer-aware span construction and score
    propagation. The lower panel groups continuation subtokens into a lexical
    span using tokenizer boundary markers. At selected layers, the upper panel
    replaces every constituent score by the maximum score within that span
    (0.24 in this example). This operation encourages joint retention under
    token-level Top-\(K\), but does not enforce the span as an indivisible
    selection unit.}
    \label{fig:span_completion_overview}
\end{figure}

\subsection{Tokenizer-Defined Lexical Spans}
\label{sec:tokenizer_span_construction}

Let
\[
x_{1:T}=(x_1,\ldots,x_T)
\]
be the token IDs and let \(p_i\) be the vocabulary piece associated with
\(x_i\). We scan the pieces from left to right. Token \(i\) starts a new span
when any of the following conditions holds. For GPT-2-style byte-level BPE,
the reversible byte-to-Unicode mapping represents the ASCII space, line-feed,
and horizontal-tab bytes by the printable vocabulary symbols
\texttt{\.{G}} (\texttt{U+0120}), \texttt{\.{C}} (\texttt{U+010A}), and
\texttt{\^{c}} (\texttt{U+0109}), respectively. These code points describe
the tokenizer's printable representations, rather than the Unicode code
points of the original whitespace characters.

\begin{itemize}
    \item \(i=1\), or \(x_i\) is a special token;
    \item \(p_i\) begins with a tokenizer-specific whitespace-boundary
    symbol: \texttt{\.{G}} for GPT-2-style byte-level BPE or the
    SentencePiece whitespace meta-symbol \texttt{U+2581};
    \item for GPT-2-style byte-level BPE, \(p_i\) begins with the printable
    representation of a line-feed or horizontal-tab byte, namely
    \texttt{\.{C}} or \texttt{\^{c}};
    \item the preceding piece \(p_{i-1}\) contains a line-feed
    or tab marker anywhere, or a space marker after its first character.
\end{itemize}

Otherwise, token \(i\) is assigned to the same span as token \(i-1\).
Consequently, BPE or SentencePiece continuation tokens without an intervening
boundary remain in one lexical span. We denote the resulting span ID by
\[
g(i)\in\{1,\ldots,G\}.
\]
This mapping depends only on the tokenizer output; it is independent of the
question, answer string, attention weights, and task labels.

For readability, let \texttt{<WB>} denote either tokenizer word-boundary
marker. For example,
\begin{quote}
\small\ttfamily
[<WB>password, :, <WB>A7, xQ, 91, Lm, Z]
\end{quote}
produces the conceptual spans
\begin{quote}
\small\ttfamily
[<WB>password, :]\qquad [<WB>A7, xQ, 91, Lm, Z].
\end{quote}
In contrast,
\texttt{[<WB>New, <WB>York, <WB>City]}
forms three spans. Thus, the procedure detects tokenizer-defined,
approximately whitespace-delimited lexical units rather than complete named
entities.

\begin{algorithm}[t]
\small
\caption{Selective Span Propagation}
\label{alg:tokenizer_span_propagation}
\begin{algorithmic}[1]
\Require Token IDs \(x_{1:T}\), pieces \(p_{1:T}\), scores \(s\),
propagation layers \(\mathcal{L}_{\mathrm{span}}\)
\State \(g\gets-1\)
\For{\(i=1,\ldots,T\)}
    \State \(\mathrm{new}\gets(i=1)\lor\operatorname{Special}(x_i)\)
    \State \(\mathrm{new}\gets\mathrm{new}\lor
    \operatorname{StartsBoundary}(p_i)\)
    \State \(\mathrm{new}\gets\mathrm{new}\lor
    (i>1\land\operatorname{ContainsBoundary}(p_{i-1}))\)
    \If{\(\mathrm{new}\)}
        \State \(g\gets g+1\)
    \EndIf
    \State \(\mathrm{span\_id}[i]\gets g\)
\EndFor
\State \(\widetilde{s}\gets s\)
\ForAll{\(\ell\in\mathcal{L}_{\mathrm{span}}\) and KV heads \(h\)}
    \ForAll{spans \(g\)}
        \State \(m_{\ell,h,g}\gets
        \max\{s_{\ell,h,i}:\mathrm{span\_id}[i]=g\}\)
    \EndFor
    \For{\(i=1,\ldots,T\)}
        \State \(\widetilde{s}_{\ell,h,i}\gets
        m_{\ell,h,\mathrm{span\_id}[i]}\)
    \EndFor
\EndFor
\State Keep the first four tokens and select the top \(K-4\) of the remaining tokens by \(\widetilde{s}\).
\end{algorithmic}
\end{algorithm}

\begin{figure*}[!t]
    \centering
    \includegraphics[width=\textwidth]{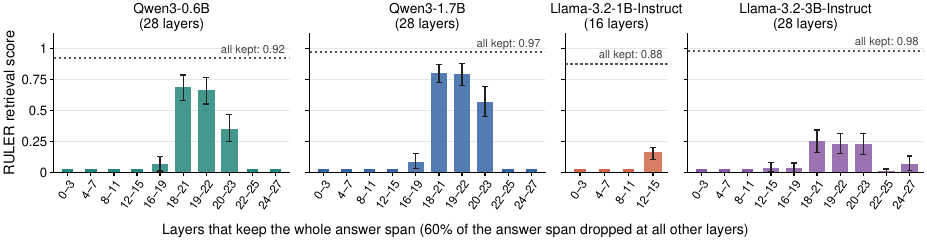}
    \caption{Effect of answer-span retention on RULER retrieval.
    At each layer outside the indicated four-layer band, we randomly evict 60\%
    of the answer tokens from the KV cache. Within the band, we keep the entire
    answer span and instead evict the same number of non-answer tokens. Dotted lines show scores
    with the answer span retained at every layer. Error bars show 95\% confidence
    intervals from 10,000 bootstrap replicates stratified by task.}
    \label{fig:layer_band_needle_retention}
\end{figure*}

\subsection{Score Update in Selective Span Propagation}
\label{sec:runtime_span_integration}

Restoring the layer and KV-head indices, let \(s_{\ell,h,i}\) denote the
original FAS-KV importance score of cached token \(i\) at layer \(\ell\) and
KV head \(h\). Let \(G(i)=\{j:g(j)=g(i)\}\) be the tokenizer-defined lexical
span containing token \(i\), and let \(\mathcal{L}_{\mathrm{span}}\) be the
layer set selected by the offline calibration described below. Before
per-head Top-\(K\) selection, we replace the scores by
\begin{equation}
\widetilde{s}_{\ell,h,i}
=
\begin{cases}
\displaystyle
\max_{j\in G(i)}s_{\ell,h,j},
& \ell\in\mathcal{L}_{\mathrm{span}},
\\[5pt]
s_{\ell,h,i},
& \ell\notin\mathcal{L}_{\mathrm{span}}.
\end{cases}
\label{eq:appendix_span_score_propagation}
\end{equation}
Thus, at each calibrated layer, every subword in a lexical span receives the
largest original score within that span; scores at all other layers remain
unchanged. At each layer, each KV head keeps the first four
tokens and the \(K-4\) highest-scoring tokens among the remaining positions,
using \(\widetilde{s}_{\ell,h,i}\).

This propagation encourages joint retention of related subwords but is not
hard group selection. If the Top-\(K\) boundary cuts through tokens tied by
Eq.~\eqref{eq:appendix_span_score_propagation}, only part of the span may still
survive. The procedure therefore does not guarantee complete span retention.
It only performs a span-wise maximum and token-wise lookup and requires no
additional attention computation.

\subsection{Motivating Layer-Band Intervention}
\label{sec:layer_band_motivation}

Propagating scores within spans at every layer is not necessarily beneficial.
Because all subtokens in a detected span inherit its maximum score, uniform
propagation can promote weakly relevant subtokens and consume the limited KV
budget, displacing independently important tokens. We therefore ask whether
span integrity is required only at particular depths.

For this diagnostic, we use 50 examples with 4,096-token contexts, ten from each of
five RULER needle-in-a-haystack tasks (three multi-key, one multi-value, and
one multi-query). After prefilling, we modify the KV cache directly and measure
the fraction of target answers found in the generated output. When we evict
60\% of the answer tokens from the KV cache at every layer, all four models score zero.

In Figure~\ref{fig:layer_band_needle_retention},
both Qwen models achieve their highest retrieval scores in two overlapping
bands, layers 18--21 and 19--22. Llama-3.2-3B-Instruct shows a smaller
increase across the bands spanning layers 18--23. For the 16-layer
Llama-3.2-1B-Instruct, the highest score occurs in the final band
(layers 12--15). This variation motivates selecting the SSP layers
\(\mathcal{L}_{\mathrm{span}}\) separately for each model.

\lstdefinestyle{calibration}{
  basicstyle=\ttfamily\fontsize{8}{9}\selectfont,
  breaklines=true,
  breakatwhitespace=false,
  columns=fullflexible,
  keepspaces=true,
  showstringspaces=false,
  frame=none,
  aboveskip=0pt,
  belowskip=0pt,
  xleftmargin=0pt,
  xrightmargin=0pt
}

\tcbset{
  calibbox/.style={
    enhanced,
    colback=gray!4,
    colframe=black!45,
    boxrule=0.5pt,
    arc=1mm,
    left=1.5mm,
    right=1.5mm,
    top=1mm,
    bottom=1mm,
    before skip=1mm,
    after skip=1mm
  }
}

\begin{figure*}[!t]
\centering
\begin{minipage}[t]{0.49\textwidth}
\vspace{0pt}
\begin{tcolorbox}[
  calibbox,
  title=\textbf{Fixed calibration context},
  colbacktitle=gray!15,
  coltitle=black,
  fonttitle=\small
]
\begin{lstlisting}[style=calibration]
NEXUS value ALPHA has code "cobaltarc731".
EMBER value GAMMA has code "lumenfold928".
PRISM value BETA has code "silvermesh562".
EMBER value ZETA has code "harborwave184".
LUMEN value DELTA has code "copperfield729".
EMBER value ALPHA has code "embervex314".
ORBIT value GAMMA has code "meadowgrid846".
EMBER value ETA has code "crystalpath205".
NEXUS value BETA has code "velvetarc490".
EMBER value DELTA has code "velvetgrid613".
PRISM value ALPHA has code "quartzmesh357".
LUMEN value BETA has code "silverlane802".
EMBER value BETA has code "quartzlane407".
ORBIT value DELTA has code "cobaltfield691".
NEXUS value GAMMA has code "harborgrid538".
EMBER value THETA has code "copperwave976".
PRISM value DELTA has code "crystalmesh143".
LUMEN value ALPHA has code "meadowlane624".
ORBIT value BETA has code "velvetpath815".
EMBER value EPSILON has code "silverfield269".
\end{lstlisting}
\end{tcolorbox}
\end{minipage}\hfill
\begin{minipage}[t]{0.49\textwidth}
\vspace{0pt}
\begin{tcolorbox}[
  calibbox,
  title=\textbf{Aggregate retrieval},
  colbacktitle=gray!15,
  coltitle=black,
  fonttitle=\small
]
\begin{lstlisting}[style=calibration]
Question:
What are the codes for EMBER values
GAMMA, ALPHA, DELTA, and BETA?

Answer with only the four codes in
that order, separated by commas.

Answer:
lumenfold928, embervex314,
velvetgrid613, quartzlane407
\end{lstlisting}
\end{tcolorbox}
\begin{tcolorbox}[
  calibbox,
  title=\textbf{Individual retrieval},
  colbacktitle=gray!15,
  coltitle=black,
  fonttitle=\small
]
\begin{lstlisting}[style=calibration]
Q: Code for EMBER value GAMMA?
A: lumenfold928

Q: Code for EMBER value ALPHA?
A: embervex314

Q: Code for EMBER value DELTA?
A: velvetgrid613

Q: Code for EMBER value BETA?
A: quartzlane407
\end{lstlisting}
\end{tcolorbox}
\end{minipage}

\caption{The single synthetic sample used to identify
\(\mathcal{L}_{\mathrm{span}}\). The known answer strings are used only for the
offline layer-intervention search; runtime spans are constructed solely from
tokenizer boundaries.}
\label{fig:span_layer_calibration_example}
\end{figure*}

\subsection{Single-Sample Layer Calibration}
\label{sec:span_layer_calibration}

We select \(\mathcal{L}_{\mathrm{span}}\) once per model using a single short
synthetic \texttt{multivalue\_uniform} sample, shown in
Fig.~\ref{fig:span_layer_calibration_example}. The fixed context contains
20 key--value facts and approximately 200 tokens. It is paired with one
aggregate query and four individual queries for the codes of EMBER values
GAMMA, ALPHA, DELTA, and BETA. No downstream benchmark data are used.

The offline intervention uses the known answer spans only to measure layer
sensitivity; this ground-truth information is not used by runtime span
construction. Outside the candidate layer set \(\mathcal{S}\),
we protect the first subword of each answer code and mask 80\% of the
answer subwords in the KV cache. At each layer in \(\mathcal{S}\), we keep all
answer subwords and mask the same number of non-answer tokens.

We score each candidate layer set on five queries using teacher forcing.
A query counts as correct only if the model predicts every masked answer
subword correctly given its gold prefix.

\textbf{Stage 1.} We select the contiguous layer interval with the highest
query accuracy, breaking ties in favor of the shortest interval.

\textbf{Stage 2.} Within that interval, we select the subset with the highest
query accuracy, breaking ties in favor of fewer layers. We reuse this subset
for all inference inputs.

Table~\ref{tab:effective_span_layers} lists the selected intervals and subsets.

\begin{table}[t]
\centering
\scriptsize
\setlength{\tabcolsep}{2pt}
\begin{tabular}{lccc}
\toprule
\textbf{Model} &
\textbf{\# Layers} &
\textbf{Stage 1} &
\textbf{Stage 2} \\
\midrule
Qwen3-1.7B & 28 & $[11,21]$ & $\{11,16,19,21\}$ \\
Llama-3.2-1B-Instruct & 16 & $[5,14]$ & $\{5,6,8,10,13,14\}$ \\
Qwen3-8B & 36 & $[19,31]$ & $\{19,22,23,24,29,31\}$ \\
Llama-3.1-8B-Instruct & 32 & $[16,24]$ & $\{16,18,22,24\}$ \\
\bottomrule
\end{tabular}
\caption{Effective span-propagation layers identified by the two-stage
single-sample intervention search. Stage 1 selects a contiguous range and
Stage 2 selects a sparse subset. Layer indices are zero-based.}
\label{tab:effective_span_layers}
\end{table}

\subsection{Complexity}

Span IDs are constructed once per input in \(\mathcal{O}(T)\) time. At each
selected layer, the span-wise scatter-max and token-wise gather require
\(\mathcal{O}(H_{\mathrm{KV}}T)\) operations. The total propagation overhead is
therefore
\[
\mathcal{O}
\left(
|\mathcal{L}_{\mathrm{span}}|H_{\mathrm{KV}}T
\right),
\]
with no additional attention-matrix computation.

\makeatletter
\setlength{\@fptop}{0pt}
\setlength{\@dblfptop}{0pt}
\setlength{\@dblfpsep}{18pt}
\makeatother

\begin{figure*}[!t]
  \centering
  \includegraphics[width=\textwidth]{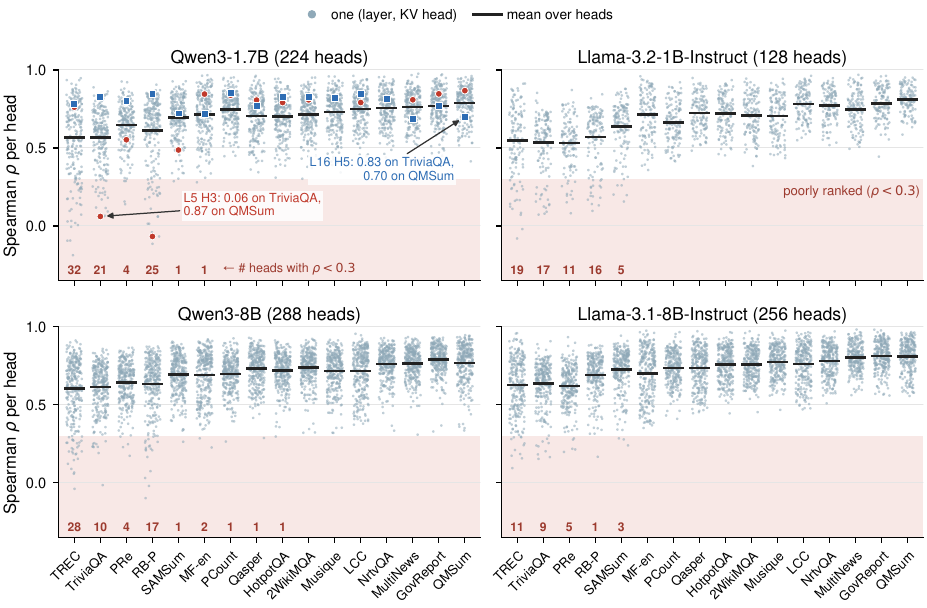}
  \caption{Per-head prediction quality on held-out LongBench inputs (10 per task, 4K context). Each dot represents one layer and KV head on one task. Spearman's $\rho$ compares the ranking of cached tokens by $w^\top k_i$ with their ranking by realized future attention $\alpha_i^{\mathrm{future}}$. Black bars mark per-task means across heads. Red counts give the number of heads below $\rho=0.3$.}
  \label{fig:prediction_errors}
\end{figure*}

\section{AIME Results}
\label{app:aime_results}

We evaluate mathematical reasoning on AIME 2024~\citep{aime2024} and
AIME 2025 \mbox{\citep{aime2025}} (30 problems each).\footnote{\url{https://huggingface.co/datasets/HuggingFaceH4/aime_2024} and \url{https://huggingface.co/datasets/alessiodevoto/aime25}.}
We use Qwen3-8B in thinking mode with its chat template in a zero-shot setting.
We use stochastic decoding with temperature $0.6$, top-$p=0.95$, and
top-$k=20$. Generation terminates at an EOS token or after 32,768 new tokens. For
compressed runs, we reduce the KV cache to 2,048 tokens every 128 decoding
steps, giving a maximum steady-state size of approximately 2,175 tokens
between eviction events.

We evaluate Full KV, FAS-KV, FAS-KV + SSP, Expected Attention, and KeyDiff.
Each configuration is run with three sampling seeds (42, 43, and 44), and we
report mean accuracy and sample standard deviation. Experiments use
FlashAttention-2 on a single NVIDIA H200 GPU. Table~\ref{tab:aime_results}
shows that SSP improves the core FAS-KV scorer by 4.5 points on AIME24 and 6.7
points on AIME25. FAS-KV + SSP matches Expected Attention on AIME24 and
outperforms it on AIME25. On AIME24, SSP also reduces the sample standard
deviation from 5.1 to 1.9 points.

SSP is not specialized to numbers. It propagates the maximum FAS-KV score
within each tokenizer-defined, approximately whitespace-delimited lexical span
as described in Sections~\ref{sec:tokenizer_span_construction}
and~\ref{sec:runtime_span_integration}. During repeated decoding-time
compression, span IDs are updated to cover both prompt and newly generated
tokens. Thus, SSP can reduce partial eviction of multi-token numerals,
identifiers, expressions, and intermediate results generated during reasoning.
This provides a plausible explanation for the AIME gains, although the
experiment does not isolate these token types causally.
This is a single-model mathematical-reasoning evaluation with repeated eviction at a fixed cache budget; it should not be interpreted as evidence of robustness on streaming, long-horizon generation, or interactive and multi-turn workloads.

\begin{table}[t]
\centering
\small
\setlength{\tabcolsep}{5pt}
\begin{tabular}{lcc}
\toprule
\textbf{Method} & \textbf{AIME24} & \textbf{AIME25} \\
\midrule
Full KV & $73.3 \pm 3.3$ & $65.6 \pm 5.1$ \\
\hdashline
FAS-KV & $\underline{31.1 \pm 5.1}$ & $\underline{24.4 \pm 1.9}$ \\
FAS-KV + SSP & $\mathbf{35.6 \pm 1.9}$ &
$\mathbf{31.1 \pm 1.9}$ \\
Expected Attention & $\mathbf{35.6 \pm 5.1}$ & $18.9 \pm 1.9$ \\
KeyDiff & $17.8 \pm 5.1$ & $20.0 \pm 0.0$ \\
\bottomrule
\end{tabular}
\caption{AIME accuracy with Qwen3-8B in thinking mode at a target KV-cache
budget of 2,048 tokens. Values are mean $\pm$ sample standard deviation over
three runs. Bold indicates the best compressed result in each column,
including ties, and underlining indicates the second-best result; Full KV is
shown separately as an upper bound.}
\label{tab:aime_results}
\end{table}

\begin{figure*}[!t]
  \centering
  \includegraphics[width=\textwidth]{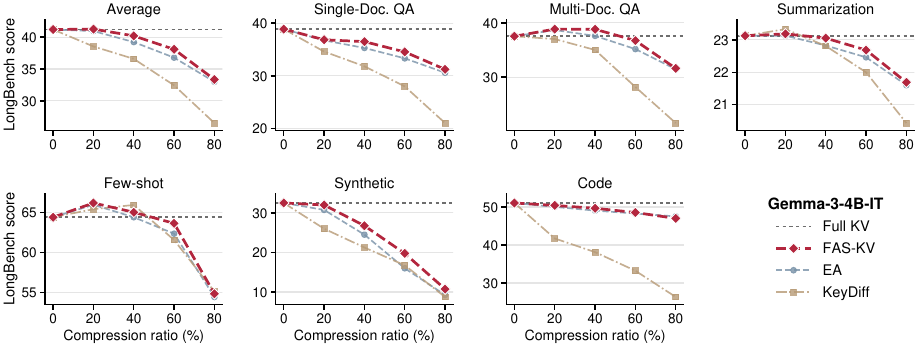}
  \caption{LongBench category-level performance of FAS-KV on Gemma-3-4B-IT
  under increasing KV-cache compression. Panels show the overall average
  and results for single-document QA, multi-document QA, summarization,
  few-shot learning, synthetic tasks, and code tasks. The results support the
  effectiveness of FAS-KV beyond the Qwen and Llama model families.}
  \label{fig:longbench_gemma3_results}
\end{figure*}

\begin{figure*}[!t]
  \centering
  \includegraphics[width=\textwidth]{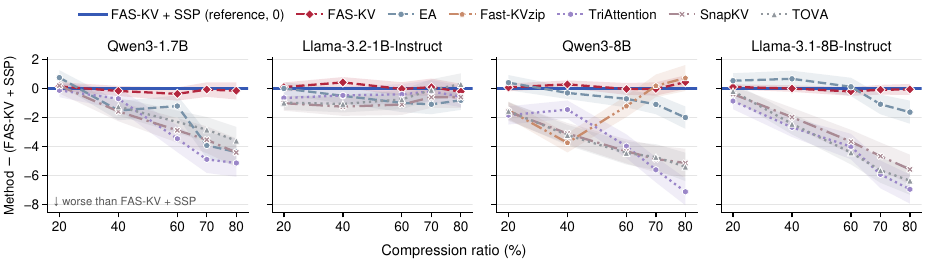}
  \caption{LongBench score differences relative to FAS-KV + SSP for four models. Negative values favor FAS-KV + SSP.
  For each point, we resample examples with replacement within each of the 16 tasks, use the same draws across methods, and average the task differences equally.
  Shading shows pointwise 95\% confidence intervals given by the 2.5th and 97.5th percentiles of 10,000 paired bootstrap replicates.}
  \label{fig:longbench_bootstrap_diff}
\end{figure*}

\begin{figure*}[!t]
  \centering
  \includegraphics[width=\textwidth]{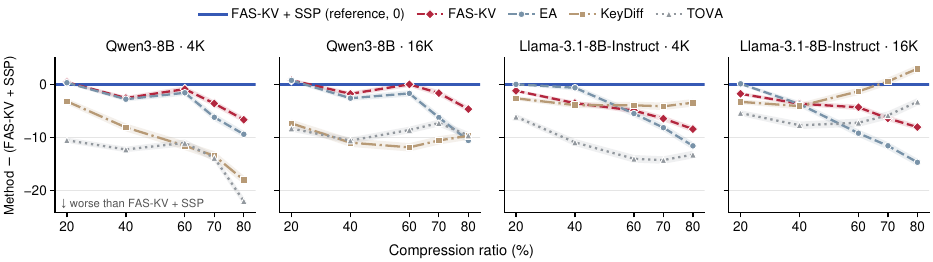}
  \caption{RULER score differences relative to FAS-KV + SSP at 4K and 16K context lengths. Negative values favor FAS-KV + SSP.
  For each point, we resample examples with replacement within each of the 13 tasks, use the same draws across methods, and average the task differences equally.
  Shading shows pointwise 95\% confidence intervals given by the 2.5th and 97.5th percentiles of 10,000 paired bootstrap replicates.}
  \label{fig:ruler_bootstrap_diff}
\end{figure*}

\section{Detailed LongBench and RULER Results}
\label{app:detailed_benchmark_results}

\begin{figure*}[p]
  \centering
  \includegraphics[width=\textwidth]{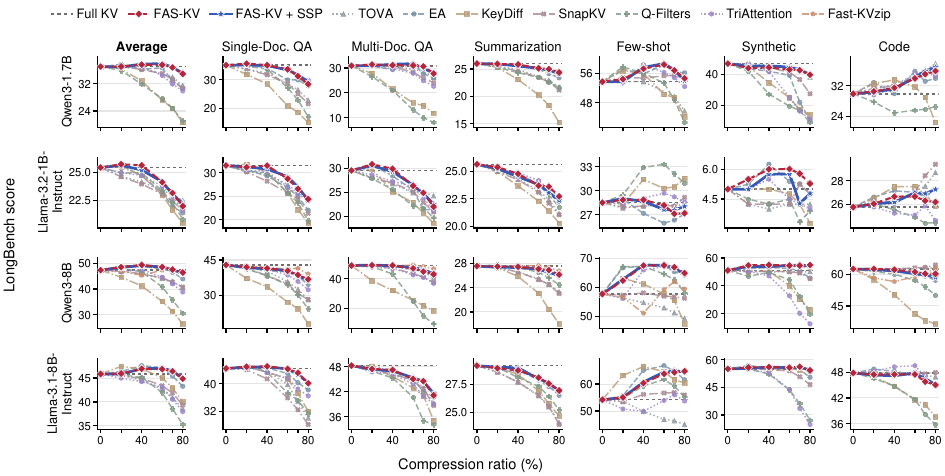}
  \caption{LongBench category-level performance under increasing KV-cache
  compression. Rows correspond to Qwen3-1.7B, Llama-3.2-1B-Instruct,
  Qwen3-8B, and Llama-3.1-8B-Instruct. Columns report the overall average and
  results for single-document QA, multi-document QA, summarization, few-shot
  learning, synthetic tasks, and code tasks.}
  \label{fig:longbench_category_results}
\end{figure*}

\begin{figure*}[p]
  \centering
  \includegraphics[width=\textwidth]{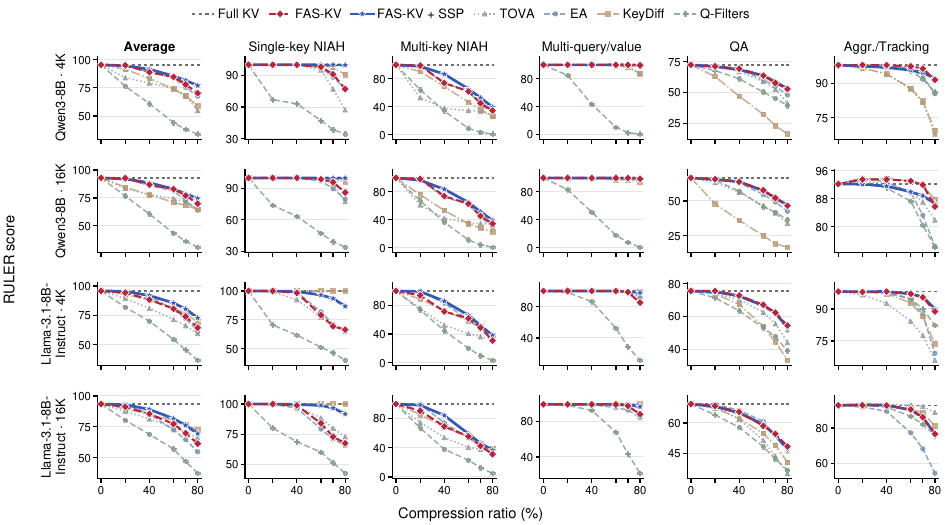}
  \caption{RULER category-level performance under increasing KV-cache
  compression. Rows correspond to Qwen3-8B and Llama-3.1-8B-Instruct at 4K
  and 16K context lengths. Columns report the overall average and results for
  single-key NIAH, multi-key NIAH, multi-query/value retrieval, QA, and
  aggregation/tracking tasks.}
  \label{fig:ruler_category_results}
\end{figure*}

Figures~\ref{fig:longbench_category_results} and
\ref{fig:ruler_category_results} expand the aggregate results in
Figures~\ref{fig:longbench_results} and~\ref{fig:ruler_results}, respectively.
Figure~\ref{fig:longbench_gemma3_results} additionally evaluates
Gemma-3-4B-IT~\citep{gemmateam2025gemma3}. FAS-KV remains competitive with EA
and substantially improves over KeyDiff in the overall average, QA, and code
categories, supporting its effectiveness beyond the Qwen and Llama model
families.

\subsection{Prediction Errors by Task and KV Head}
\label{app:prediction_errors}

Across the four models, task means range from 0.53 to 0.81, yet most heads with $\rho<0.3$ occur on a few tasks, including TREC, TriviaQA, and RepoBench-P (Figure~\ref{fig:prediction_errors}). Within each model family, the 8B model has fewer task/head pairs below this threshold than the smaller model. Llama-3.1-8B-Instruct has the fewest overall. On Qwen3-8B, 28 heads still fall below the threshold on TREC and 17 on RepoBench-P. In Qwen3-1.7B, which task yields the higher rank correlation depends on the head. Layer~5, head~3 scores higher on QMSum than on TriviaQA (0.87 vs.\ 0.06), whereas layer~16, head~5 scores higher on TriviaQA (0.83 vs.\ 0.70). How these ranking errors affect downstream accuracy or vary with context length and retrieval position remains unclear.

\clearpage

\end{document}